\documentclass[10pt,twocolumn,letterpaper]{article}

\usepackage{floatrow}

\usepackage{iccv}
\usepackage{times}
\usepackage{epsfig}
\usepackage{graphicx}
\usepackage{amsmath}
\usepackage{amssymb}
\usepackage[utf8]{inputenc}
\usepackage{mathtools}

\usepackage{algorithm}
\usepackage{algpseudocode}
\usepackage{caption}
\usepackage{subcaption}
\usepackage{xcolor}
\usepackage{pgfplots}
\usepackage{tikz}

\usepackage[font=footnotesize,skip=0pt]{caption}

\usepackage[pagebackref=true,breaklinks=true,letterpaper=true,colorlinks,bookmarks=false]{hyperref}

\iccvfinalcopy 

\def\iccvPaperID{****} 

\ificcvfinal\fi

\definecolor{bblue}{HTML}{4F81BD}
\definecolor{rred}{HTML}{C0504D}
\definecolor{ggreen}{HTML}{9BBB59}
\definecolor{ppurple}{HTML}{9F4C7C}

\begin{document}

\title{Taxonomic Class Incremental Learning}


\author{
Yuzhao Chen\thanks{The authors contribute equally to this paper.}~~~ 
Zonghuan Li\footnotemark[1]~~~ 
Zhiyuan Hu~~~ 
Nuno Vasconcelos~~~\\
\smallskip 
\\
UC San Diego~~~
\smallskip
\\
\{\tt\small{yuc103, zol005, z8hu, nvasconcelos\}@ucsd.edu}}



\maketitle
\ificcvfinal\thispagestyle{empty}\fi

\begin{abstract}
   The problem of continual learning has attracted rising attention in recent years. However, few works have questioned the commonly used learning setup, based  on a task curriculum of random class. This differs significantly from human continual learning, which is guided by taxonomic curricula. In this work, we propose the \textbf{Taxonomic Class Incremental Learning (TCIL)} problem. In TCIL, the task sequence is organized based on a taxonomic class tree. We unify existing approaches to CIL and taxonomic learning as parameter inheritance schemes and introduce a new such scheme for the TCIL learning. This enables the incremental transfer of knowledge from ancestor to descendant class of a class taxonomy through parameter inheritance. Experiments on CIFAR-100 and ImageNet-100 show the effectiveness of the proposed TCIL method, which outperforms existing SOTA methods by 2\% in terms of final accuracy on CIFAR-100 and 3\% on ImageNet-100.
\end{abstract}

\section{Introduction}

For humans, the process of learning {\it continually} is quite natural. New classes are easily integrated with existing ones, without the need to revisit data originally used to learn what is already known. This is unlike deep learning models, which fare poorly in the continual learning setting and are prone to catastrophic forgetting~\cite{lwf}: as it is trained on data from new classes, the classifier tends to forget those previously learned. This has motivated interest in continual learning, most notably the {\it class incremental problem\/} (CIL). In this setting, a  sequence of classification tasks is defined, where each task includes a set of classes that do not overlap with those of the previous tasks. At each learning step, the classifier only has access to data of the current task but is expected to remember how to classify all previously learned classes. While extensive research has been devoted to the CIL problem~\cite{lwf, icarl, der,nscl, dytox, survey1, survey2, survey3}, incremental learning has shown to be quite challenging. Current state of the art (SOTA) methods cannot match the performance of the joint classifier (all tasks learned simultaneously) even on simple datasets like CIFAR-100. This is at odds with the apparent simplicity of incremental learning for humans.

\begin{figure} \RawFloats
\begin{center}
    \includegraphics[width=\linewidth]{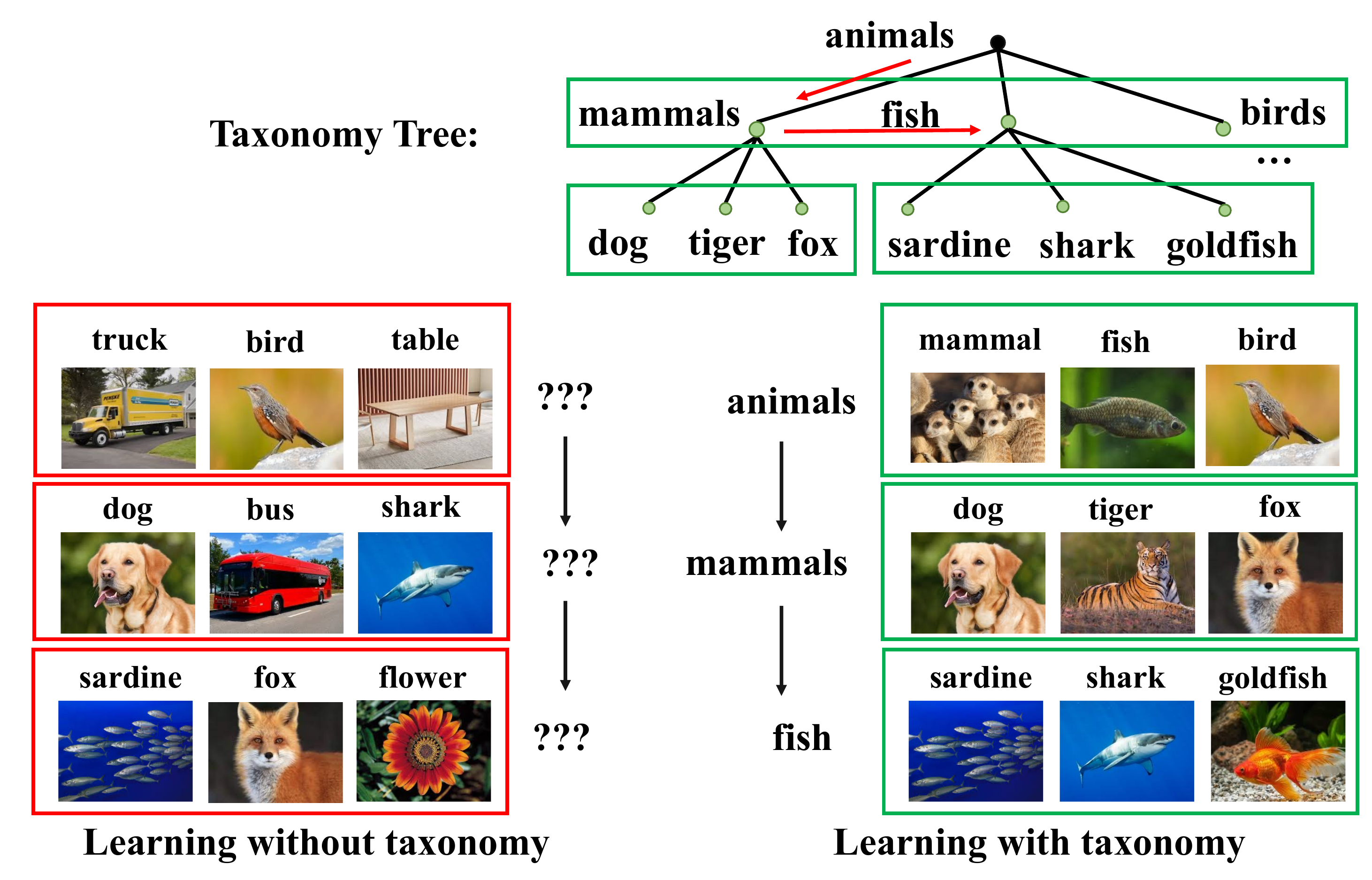}
    \caption{CIL learns using a curriculum of randomly organized tasks (red categories), which ignores the latent hierarchical class structure across classes. This is unlike most human learning, which is usually driven by a class taxonomy. TCIL replicates this setting, leveraging a class taxonomy in two ways. First, it uses a taxonomic curriculum (green categories), where classes are introduced in a coarse-to-fine manner, Second, it uses a taxonomic classifier, whose parameters capture taxonomic constraints through parameter inheritance relations.} \label{fig:teaser}
\end{center}
\end{figure}

In this work, we pose the hypothesis that this difficulty is at least partly due to the differences between the human and CIL learning settings. 
In CIL, class sets are constructed by simply randomly splitting the class label set $\mathcal{Y}$ of the joint classifier. 
This, however, is very different from the continual learning setting used to teach humans. Rather than learning randomly, people typically follow a well structured curriculum to learn new tasks incrementally. 
In many cases, this curriculum is the result of decades of optimization. For example, students first learn simple mathematics, like arithmetic, and then move on to more advanced concepts, like calculus.

For perceptual problems like object recognition, these curricula rely extensively in class taxonomies. Humans start by learning to recognize coarse concepts, such as dogs vs cats, and only then progress towards learning fine-grained classes, such as dog or cat breeds. This expertise is usually acquired {\it incrementally\/}. 

Since taxonomic learning is so intrinsically connected to continual learning for humans, it appears natural to hypothesize that task curricula defined taxonomically may enhance the ability of a vision system to learn incrementally. This motivates the introduction of {\it taxonomic class incremental learning\/} (TCIL), as illustrated in Figure~\ref{fig:teaser}. This extends the CIL setting along two dimensions: 1) introduction of a {\it class taxonomy,\/} and 2) definition of CIL tasks according  to a {\it taxonomic curriculum\/} that progresses from coarse-grained to fine-grained classes. In this work, we propose a procedure to create taxomomic curricula by breadth first descent over a class taxonomy, and a procedure to derive the associated datasets from a ``flat" dataset, i.e. a dataset labeled only with respect to the class label set $\cal Y$ of the joint for the classes at the leaves of the tree, as is usual in the literature. This enables the investigation of TCIL approaches for any of the datasets commonly used in the recognition literature. We then use this procedure to create TCIL versions of the CIFAR-100 and ImageNet-100 datasets, and use these datasets to investigate TCIL methods.

We then consider how to design a classifier that leverages the hierarchical constraints inherent to a taxonomic tree. We show that a key concept for this is {\it parameter inheritance\/} between classifiers. Parameter inheritance is frequently used in the taxonomic classification literature~\cite{deeprtc} to ensure that the classifier parameters are defined in a coarse-to-fine manner as one descends the tree. We show that network expansion, the core concept of various SOTA CIL methods, is also a form of parameter inheritance, where a classifier inherits parameters from those of the previous tasks. We then unify the two forms of inheritance into a new approach to TCIL, which supports both a taxonomic curriculum and a taxonomic classifier.   

We finally compare several methods to solve the TCIL problem. We start by considering the setting where the task curriculum is derived from the taxonomy, but the architecture does not leverage the latter. This boils down to the application of the SOTA DER~\cite{der} approach for CIL to the sequence of fine-grained tasks posed by TCIL. We show that the TCIL curriculum significantly improves CIL performance. We next consider the full TCIL setting, where the model is itself a taxonomic classifier, and show that this outperforms CIL-style learning of a flat classifier under any of the curricula considered, either taxonomic or not. 

Overall, the paper makes the following contributions:
\begin{enumerate}
    \item Definition of the TCIL problem, and procedures to create TCIL datasets from existing classification datasets.
    \item Deep learning methods that leverage SOTA solutions to both CIL and taxonomic classification to produce novel models that explicitly solve the TCIL problem.
    \item An experimental evaluation showing both the benefits of semantic curricula over random curricula even for CIL, and of TCIL of a hierarchical classifier over CIL of a flat one under any type of curriculum.
\end{enumerate}

\section{Related Works}
\noindent\textbf{Incremental learning:} Incremental Learning, or Continual Learning, aims to learn a sequence of tasks without forgetting. IL assumes the data of old tasks are on longer available or can only be kept in a small memory buffer, this is fatal to most gradient based neural networks and leads to severe \textit{catastrophic forgetting} problem\cite{lwf, survey1, survey2, survey3}. \textit{Distillation based} methods\cite{lwf, icarl, podnet, foster, dytox} keep the old model trained on the last task, input data is processed in both old and new models. The new model is expected to align with old model in terms of some intermediate outputs (logits, intermediate features, etc.), this is typically implemented by a distillation loss\cite{distillation}. \textit{Parameter Consolidation based} methods\cite{ewc, gem, geodl, si, afc} recognize important parameters of current task using some importance metric(e.g. Fisher Information\cite{fisher}, PCA coefficients). Penalties will be applied if future model changes those important parameters. \textit{Nullspace Projection based} methods\cite{nscl, adns} move one step forward: once the important parameters are recognized, future model can only learn in the null space of those parameters, so that the knowledge of old tasks won't be affected. \textit{Parameter Isolation based} methods\cite{expert,der,pnn,piggy,rpsnet, wsn}, also known as \textit{Network Expansion based} methods, fix the model trained on old tasks so the knowledge won't be forgotten at all. However, to learn new tasks, those methods will have to add new sub-networks, thus leads to a continually growing model size. Some parameter isolation based methods\cite{pnn, wsn, progressprompt} add cross-connections from old networks the new network. These cross connections can effectively transfer knowledge from old to new and lead to better performances, but also further increase the model size. Although some methods\cite{piggy, wsn} try to reduce network scale by network purging, how to reach a better performance-scale trade-off is still a challenging problem.

\noindent\textbf{Learning with taxonomy:} Hierarchical structure is widely existed in various machine learning datasets and problems\cite{hier_survey}. In the domain of visual classification, widely used datasets like ImageNet\cite{imagenet}, iNaturalist\cite{iNat} are build from a taxonomy tree like WordNet\cite{wordnet}. Although most classification problem only focus on categories that are on the leaf nodes, the model with a flat classifier over leaf nodes can only reach sub-optimal results\cite{hier_survey}. To better understand the taxonomy, various methods\cite{largemargin, deeprtc, bcnn, tnn, hdcnn, expnet} have been proposed. These methods typically use a hierarchy sturcture in network design, features or logits of leaf nodes are integrated with their corresponding ancestor nodes. 

In continual learning, taxonomy is seldom addressed. This is because most continual learning methods are trained and evaluated on small scale and laboratory dataset like MNIST\cite{mnist}, CIFAR-10\cite{cifar}, CIFAR-100\cite{cifar} and ImageNet-100\cite{icarl}. However, for the continual learning of human beings in the real world, a curriculum based on taxonomy is crucial for the learning process. In this work we consider the continual learning problem that follows a taxonomic curriculum.

\section{Taxonomic Class Incremental Learning}

In this section, we formulate the problem of taxonomic class incremental learning (TCIL). 

\noindent\textbf{Class Incremental Learning:} Class incremental learning (CIL) addresses the problem of learning a sequence of $N_T$ classification tasks $\mathcal{T}=\{\mathcal{T}_1, \mathcal{T}_2, \dots, \mathcal{T}_{N_T}\}$ incrementally and without forgetting. The $i$-th task $\mathcal{T}_i$ has access to a dataset of $M_i$ samples $D_i = \{(x_j, y_j)\}_{j=1}^{M_i}$ from examples of  class or label set $\mathcal{Z}_i$. The class sets that define different tasks are disjoint, i.e. $\mathcal{Z}_i\cap \mathcal{Z}_j=\varnothing, \forall i\neq j$. While the model only has access to $D_i$ to learn task $\mathcal{T}_i$, incremental learning requires that it remembers all previous tasks after this learning. Hence, performance is evaluated on a dataset of all known classes $\mathcal{Y}_i=\mathcal{Z}_1\cup\mathcal{Z}_2\cup\dots\mathcal{Z}_i$. Since forgetting is difficult to avoid without access to any data from previous tasks, most CIL methods maintain a small memory buffer $\mathcal{B} (|\mathcal{B}| << |D_i|)$ of examples from previous tasks. The model of task $i$ is thus trained on $D_i\cup\mathcal{B}$.

\noindent\textbf{Taxonomic Class Incremental Learning: } \label{sec_task_formulation}
A taxonomic class incremental (TCIL) problem is defined with respect to a class taxonomy tree $\mathcal{H}$ of $N$ leaf node classes $\mathcal{L}(\mathcal{H})$. We denote by $\mathcal{R}(\mathcal{H})$ the set of all non-leaf nodes. Tasks are defined sequentially, by visiting each node in the tree in a breadth first manner, starting at the root node. At step $i$, a task ${\cal T}_i$ is defined by visiting node $\mathcal{N}_i \in \mathcal{R}(\mathcal{H})$, and augmenting the class set with the child nodes $\mathcal{C}(\mathcal{N}_i)$ of $\mathcal{N}_i$. This creates a new sub-tree ${\cal H}_i$ whose leaf nodes determine the label set of the task, i.e. ${\cal Y}_i = {\cal L}({\cal H}_i)$. The process terminates when the class set includes all the leaf nodes $\mathcal{L}(\mathcal{H})$. The number of total tasks is therefore $N_T = |\mathcal{R}(\mathcal{T})|$. 

\begin{figure} \RawFloats
\begin{center}
    \includegraphics[width=\linewidth]{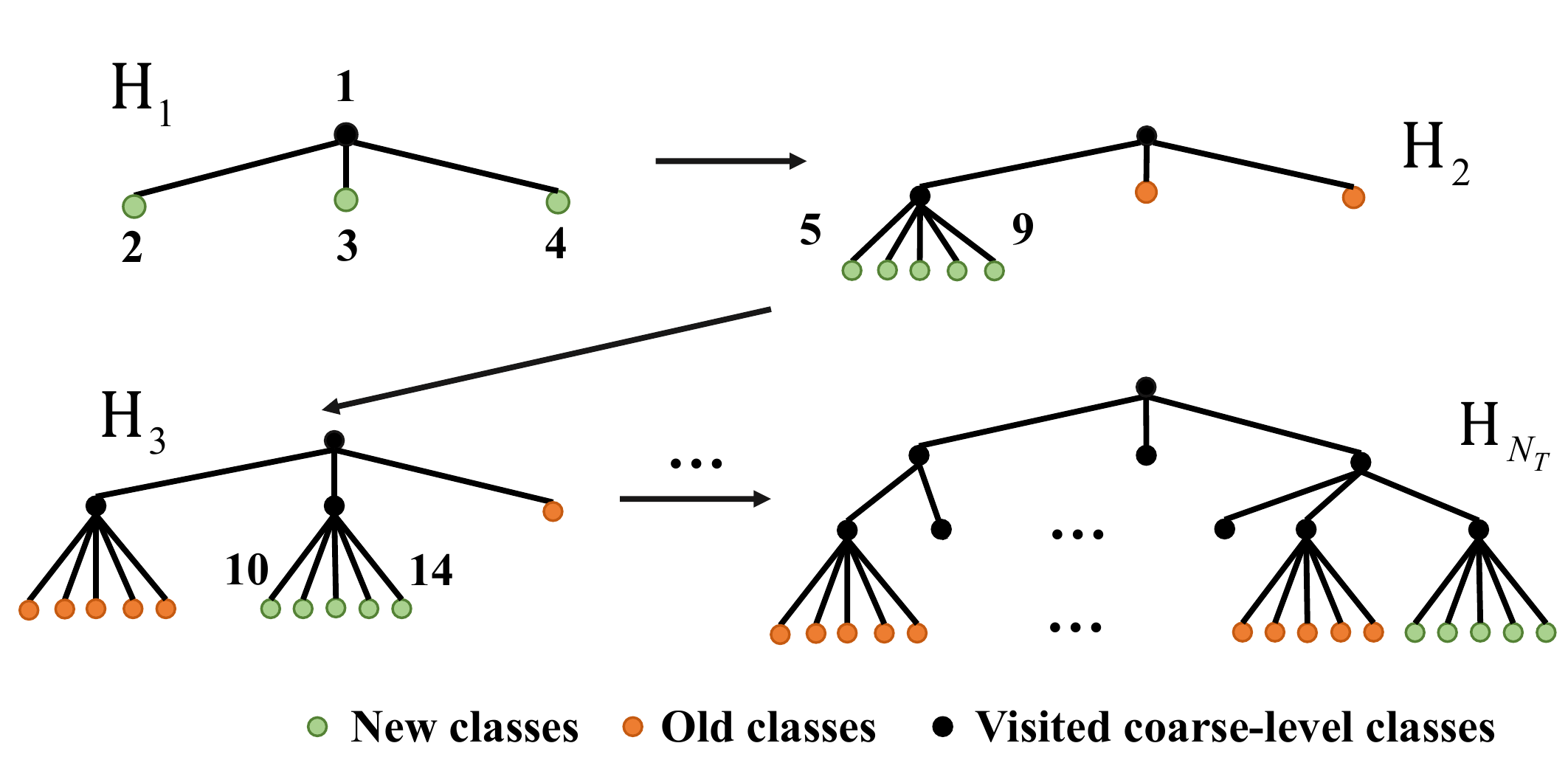}
    \caption{Example of task definition under TCIL. At step $i$, a new task ${\cal T}_i$ is defined by expanding the class set with the children of node ${\cal N}_i$. This defines a taxonomic classifier of tree ${\cal H}_i$, which includes a mix of old classes, shown in orange, and new classes, shown in green. The coarse-level classes already visited in previous tasks are shown in black.}\label{fig:tree_expand}
\end{center}
\end{figure}

The process is illustrated in Figure~\ref{fig:tree_expand}. In this example, task ${\cal T}_1$ includes the children nodes of the root, i.e. it has label set ${\cal Y}_1 = \{{\cal N}_2, \ldots, {\cal N}_4\}$ as shown at the top of the figure. Task ${\cal T}_2$ then expands the label set to include the children of ${\cal N}_2$, i.e. ${\cal Y}_2 =  \{{\cal N}_3, \ldots, {\cal N}_9\}$. This task includes two coarse-grained classes previously defined in ${\cal T}_1$, nodes $\{{\cal N}_3,{\cal N}_4\}$ shown in orange, and five new fine-grained classes, $\{{\cal N}_5,\ldots,{\cal N}_9\}$, which are shown in green. As usual in CIL, the task is learned using a dataset of the new classes only, which is denoted as ${\cal D}_2$. Task ${\cal T}_3$ then expands the class set with the children of node ${\cal N}_3$, i.e. ${\cal Y}_3 =  \{{\cal N}_4, \ldots, {\cal N}_{14}\}$, and so forth, until the last task, ${\cal T}_{N_T}$, which has all leaf nodes as class set. 

Note that, as shown on the right hand-side of the figure, each task ${\cal T}_i$ is itself a taxonomic classification of class-tree 
\begin{equation}
    \mathcal{H}_i = \bigcup_{j=1}^i (\mathcal{N}_j \cup \mathcal{C}(\mathcal{N}_j)), \label{eq:Hi}
\end{equation}
which is a subtree of ${\cal H}$. The label set of the task is then
\begin{equation}
    {\cal Y}_i = {\cal L}({\cal H}_i)
    \label{eq:yi}
\end{equation}
and the dataset ${\cal D}_i$ only contains data from classes ${\cal C}({\cal N}_i)$. In this way, subtrees ${\cal H}_i$ contain more fine-grained classes as $i$ increases, simulating the learning curriculum commonly used to teach humans incrementally, i.e. progressing from coarse to fine-grained concepts.  As is common in CIL, it is possible to include a small buffer buffer ${\cal B}$ of data from the previous classes, i.e. label set ${\cal Y}_i \setminus {\cal C}({\cal N}_i)$. Since this usually improves performance substantially, we consider it as the default setting for TCIL. While the process can be applied to any tree, in this work we only consider the case where $\cal H$ is a balanced tree, i.e. each node of the same depth in $\cal H$ has the same number of children. 

\begin{figure}\RawFloats
\begin{center}
    \includegraphics[width=\linewidth]{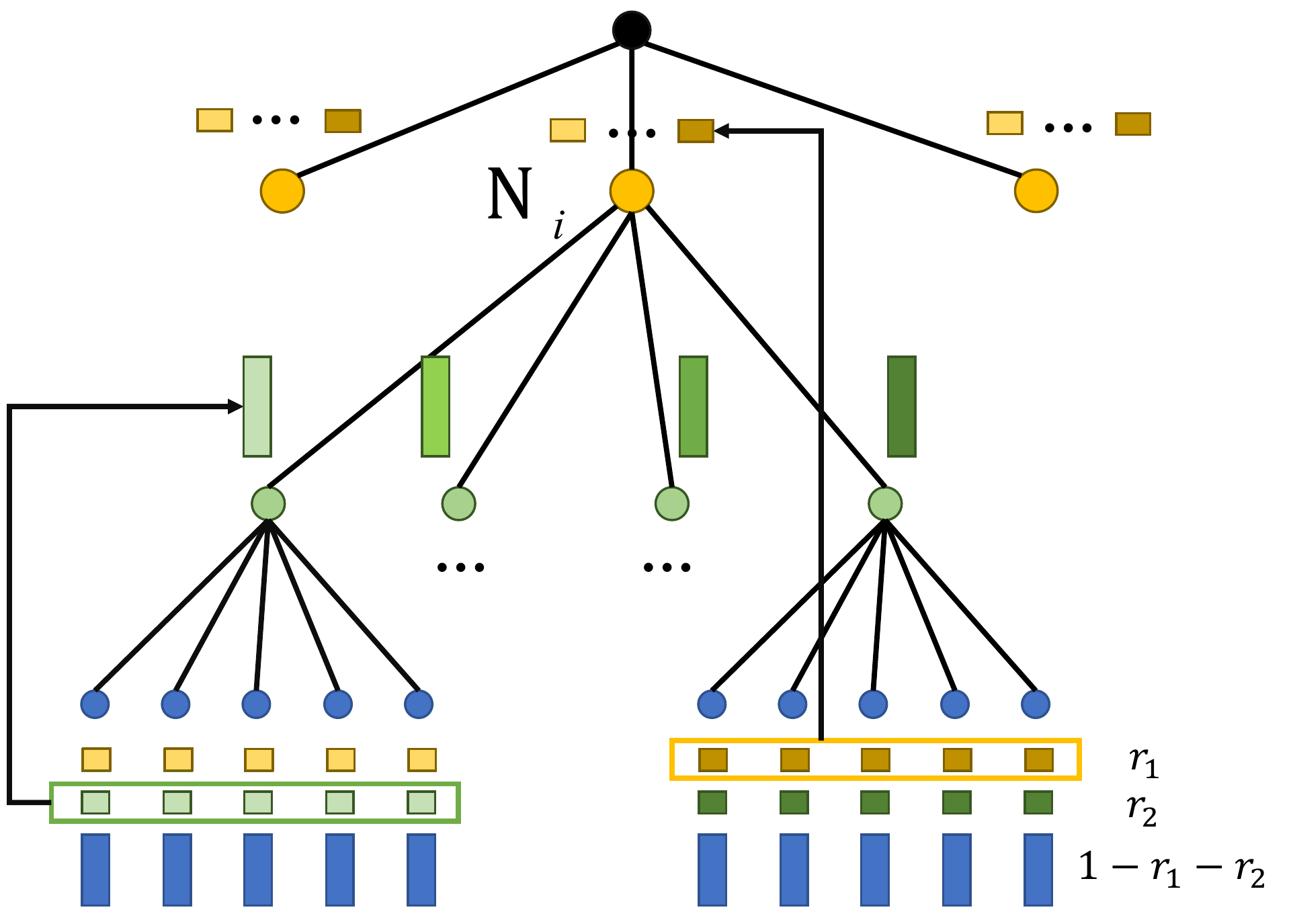}
    \caption{Example of dataset splitting. For a 3-layers taxonomic tree, datasets of coarse-level nodes are sampled from leaf nodes,  the sample rates are controlled by $r_1$ and $r_2$.} \label{fig:data_split}
\end{center}
\end{figure}

\noindent\textbf{Dataset creation:} Beyond task definition, the formulation of the TCIL problem requires the assembly of a dataset $\mathcal{D}_n$ for each task. Since the labels of most existing datasets, e.g. CIFAR or ImageNet, report to the leaf nodes, there is a need to split these datasets, to  create labeled data for the intermediate tasks ${\cal T}_i, i < N_T$. This process is illustrated in Figure~\ref{fig:data_split}, for a tree of three levels.

Given a node ${\cal N}_i$, shown in the first level of the tree, the process starts with a dataset that contains all the data of all the leaf nodes descendants of ${\cal N}_i$, i.e. the blue nodes of the figure. Defining as  ${\cal H}({\cal N})$ the subtree rooted at node ${\cal N}$, this can be written as
\begin{equation}
    {\cal D}({\cal N}_i) = \bigcup_{{\cal Y} \in {\cal L}({\cal H}({\cal N}_i))} {\cal D}({\cal Y}).
    \label{eq:Dni}
\end{equation}
where ${\cal D}({\cal Y})$ contains the data available for class ${\cal Y}$. Dataset ${\cal D}({\cal N}_i)$ includes all data shown at the bottom of the figure. For non-incremental taxonomic learning, this dataset is used to train any taxonomic classifier that involves node ${\cal N}_i$. However, the use of datasets ${\cal D}({\cal N}_i)$ for TCIL would result in repetition of the data used to train the tasks defined at coarser nodes and their descendants. In the example of the figure, ${\cal D}({\cal N}_i)$ would simply be the union of the datasets ${\cal D}({\cal N}_j)$ of the nodes ${\cal N}_j$ shown in green. Hence, any task involving the training of green nodes would be reusing data previously used to train tasks that involve node ${\cal N}_i$. This would violate the spirit of incremental learning. 

To avoid the problem, we note that the size of ${\cal D}({\cal N}_i)$ grows as one ascends the tree. In the example of the figure, each of three yellow nodes at the top of the tree would contain one third of the total data at the bottom. This is usually much larger than needed to train a classifier of the yellow nodes. Hence, the problem can be avoided by sampling data {\it without replacement,\/} i.e. randomly splitting the dataset of (\ref{eq:Dni}) into two disjoint subsets 
\begin{equation}
    {\cal D}({\cal N}_i) = {\cal S}_i \cup {\cal S}_i^c, \quad \mbox {such that } {\cal S}_i \cap {\cal S}_i^c = \emptyset
\end{equation}
where ${\cal S}_i$ contains $N_i$ examples of each of the classes in ${\cal L}({\cal H}({\cal N}_i))$. The sample ${\cal S}_i$ is the data associated with node ${\cal N}_i$. In the figure, this consists of all the yellow boxes shown next to the node. The remaining data ${\cal S}_i^c$ includes the green and blue data chunks.
This concludes the process of assembling data for the training of node ${\cal N}_i$.

The sample ${\cal S}_i$ is then removed from the dataset ${\cal D}({\cal N}_i)$ that is available for the next stage, which visits the children nodes of ${\cal N}_i$, shown in green, in succession. These nodes are considered as ${\cal N}_i$ and the process is repeated. This creates the corresponding samples ${\cal S}_i$, which are the green datasets shown next to the green nodes. Finally, the children of each green node (blue nodes) are visited, and the remaining data is used to create datasets for these nodes.

To define the sizes of the samples ${\cal S}_i$ associated with each node ${\cal N}_i$, we define a set of {\it sampling rate\/} parameters. All nodes of the same tree depth receive the same number of samples. 
For a tree $\cal H$ of depth $D$, there are $D-1$ such parameters $r_1, \dots, r_{D-1} \in (0,1)$. These satisfy $0 < \sum_{d=1}^{D-1} r_d < 1$, and the size of the sample ${\cal S}_i$ of any node ${\cal N}_i$ at depth $d$ is then $r_d N$, where $N$ is the size of the entire dataset.

Finally, given the samples ${\cal S}_j$ associated with nodes ${\cal N}_j$, the dataset of task ${\cal T}_i$ is the union of the samples at the leafs of the associated tree ${\cal H}_i$, i.e.
\begin{equation}
    {\cal D}_i = \left (\bigcup_{j | {\cal N}_j \in {\cal C}({\cal N}_i)} {\cal S}_j\right)\bigcup {\cal B}.
\end{equation} 
where ${\cal B}$ is a small buffer of data from previous tasks.

\section{Methods}

In  this section we propose a deep learning architecture for TCIL.

\noindent\textbf{Image classification:} 

As usual in deep learning, we consider a classifier that maps a space of examples $\cal X$ into a label or class set $\cal Y$, using an architecture composed by a feature extractor $\Phi({\bf x}): {\cal X} \rightarrow \mathbb{R}^d$ and a linear classifier of parameter matrix $\mathbf{M} \in \mathbb{R}^{|{\mathcal Y}|\times d}$, where $d$ is the feature space dimension. The classifier produces an estimate of the posterior probability distribution $\mathbf{\pi}(\mathbf{x})$ of the class label $y$ given image $\mathbf{x}$ using
\begin{equation}
\mathbf{\pi}(\mathbf{x}) = \rho(\mathbf{\ell}(\mathbf{x})), \quad \mathbf{\ell}(\mathbf{x}) = \mathbf{M} \Phi({\bf x}) \label{eq:pi}
\end{equation}
where $\rho(\cdot)$ is the softmax function,  $\mathbf{\ell}(\mathbf{x})$ a vector of logits, and $\pi_k({\bf x})$ an estimate of $P_{Y|X}(k|\mathbf{x})$.

\noindent\textbf{Network expansion: }
Network expansion techniques, such as WSN~\cite{wsn} or DER~\cite{der}, are known to achieve state of the art performance in the CIL setting. To learn task ${\cal T}_t$, the CIL model learned in previous tasks is expanded with a new feature extractor $\mathbf{\Phi}_t(\mathbf{x}) : {\cal X} \rightarrow \mathbb{R}^{\delta_i}$. The feature extractors $\{\mathbf{\Phi}_i\}_{i=1}^{t-1}$ learned in the previous tasks are frozen and the features are concatenated 
into a joint feature space  $\mathbf{\Phi}^{CIL}_t(\mathbf{x}) = \oplus_{i=1}^t \mathbf{\Phi}_t(\mathbf{x}) \in \mathbb{R}^{d_t}$, where $\oplus$ denotes concatenation and $d_t = \sum_{i=1}^t \delta_i$ is the joint feature dimension for task $t$. A new classifier matrix $\mathbf{M}_t^{CIL}\in\mathbb{R}^{|\mathcal{Y}_t|\times d_t}$ is then learned and the classifier implemented with (\ref{eq:pi}). 


\noindent\textbf{Parameter inheritance:}

Consider two classifiers with matrices $\mathbf{M}_a$ and $\mathbf{M}_b$ of overlapping label sets, related by an {\it inheritance relation\/} 
\begin{equation}
\mathbf{M}_b = {\cal I}_a^b(\mathbf{M}_a) = {\bf I}_a^b \mathbf{M}_a, \quad \quad {\bf I}_a^b \in \mathbb{R}^{|{\mathcal Y}_b|\times |{\mathcal Y}_a|} 
\end{equation}
that determines how the parameters of $\mathbf{M}_a$ are reused by $\mathbf{M}_b$. For example, assuming that $|{\mathcal Y}_a| = k$, if  classes $2$ to $k$ of ${\mathcal Y}_a$ become classes $1$ to $k-1$ of ${\mathcal Y}_b$, then
 \begin{equation}
     {\bf I}_a^b = \left[\begin{array}{c}
  \mathbf{\Sigma}_k  \\ \mathbf{0}_{(|{\mathcal Y}_b|-k +1) \times k}
  \end{array} \right] 
 \quad
  {\tiny
  \mathbf{\Sigma}_k = 
  \left[\begin{array}{ccccc}
  0,& 1,& 0, & \dots, & 0 \\
  0,& 0,& 1, & \dots, & 0 \\
    &   &    & \vdots &   \\
  0,& 0,& 0 & \dots & 1  
  \end{array} \right]} 
  \label{eq:inheritance}
\end{equation}
where $\mathbf{0}_{m\times n}$ represents a zero matrix with size $m\times n$.

Under the parameter inheritance approach, the parameters of the two classifiers are then related by
\begin{equation}
    \mathbf{M}_{b} = {\bf I}_a^b \mathbf{M}_a + \mathbf{R}_b
    \label{eq:Mb}
\end{equation}
where $\mathbf{R}_b\in\mathbb{R}^{|\mathcal{Y}_b|\times d}$ is a trainable refinement matrix and $d$ is the dimension of the feature space.

While standard parameter inheritance is performed on a fixed feature space, methods like network expansion can be seen as parameter inheritance schemes performed over dynamically expanding feature spaces. This can be accommodated by introducing an expansion operator $\mathbf{E}_{d_a}^{d_b} (d_b > d_a)$ which zero pads a matrix on the right according to
\begin{align}
    \mathbf{E}_{d_a}^{d_b} &= \left[\text{Id}(d_a)\quad \mathbf{0}_{d_a\times(d_b-d_a)}\right]\in\mathbb{R}^{d_a \times d_b} \label{eq:EO}\\
    \mathbf{M}\mathbf{E}_{d_a}^{d_b} &= \left[\mathbf{M}, \, \mathbf{0}_{m \times (d_b-d_a)} \right]\in\mathbb{R}^{m \times d_b}
\end{align}
where $\text{Id}(d_a)$ is the identity matrix of size $d_a\times d_a$ and $\mathbf{M}\in\mathbb{R}^{m \times d_a}$ is the matrix of a $m$ class classifier and feature space dimension $d_a$. If $d_a=d_b$ (no feature expansion), $\mathbf{E}_{d_a}^{d_b}$ is an identity matrix. This allows the generalization of the parameter inheritance relation of (\ref{eq:Mb}) to the case where $\mathbf{M}_{b}$ has a higher dimensional feature space than $\mathbf{M}_{a}$, using
\begin{equation}\label{eq:expansionPI}
    \mathbf{M}_{b} = {\bf I}_a^b \mathbf{M}_a \mathbf{E}_{d_a}^{d_b}+ \mathbf{R}_b
\end{equation}
where $\mathbf{M}_a \in \mathbb{R}^{|{\cal Y}|_a \times d_a}$ and $\mathbf{M}_b, \mathbf{R}_b \in \mathbb{R}^{|{\cal Y}|_b \times d_b}$, $d_b \geq d_a$.

\noindent\textbf{Inheritance in CIL classifier: }
In CIL, the class label set grows from $\mathcal{Y}_{t-1}$ to $\mathcal{Y}_t$ at task $\mathcal{T}_t$. Since the old $|\mathcal{Y}_{t-1}|$ classes remain the same, the new classifier $\mathbf{M}_t^{CIL}$ inherits the classifier matrix of $\mathbf{M}_{t-1}^{CIL}$ of the previous iteration. This can be written as in  (\ref{eq:expansionPI})
with 
\begin{equation}\label{eq:expansionCIL}
    \mathbf{M}^{CIL}_{t} = {\bf I}_{t}^{CIL} \mathbf{M}^{CIL}_{t-1} \mathbf{E}_{d_t}^{d_{t-1}}+ \mathbf{R}^{CIL}_t
\end{equation}
where
\begin{equation}\label{equ:CIL_PI}
    {\bf I}_{t}^{CIL} = \left[\begin{array}{c}
    \text{Id}(|\mathcal{Y}_{t-1}|)\\
    \mathbf{0}_{(|\mathcal{Y}_t|-|\mathcal{Y}_{t-1}|)\times |\mathcal{Y}_{t-1}|} \\
  \end{array} \right]
\end{equation}
is the inheritance matrix of task $\mathcal{T}_t$. However, most CIL methods only rely on 
inheritance for parameter initialization. In step $t$, the entire matrix $\mathbf{M}^{CIL}_{t}$ is updated. Hence $\mathbf{R}^{CIL}_t$ has the structure 
\begin{equation}\label{eq:CIL_REF}
    \mathbf{R}_t^{CIL} = \left[\begin{array}{cc}
    \Delta\mathbf{M}^{CIL}_{t-1} & \mathbf{F}^t_{t-1}\\
    \mathbf{F}^{t-1}_t & \mathbf{M}_t
  \end{array} \right]
\end{equation}
where $\Delta\mathbf{M}^{TIL}_{t-1}\in\mathbb{R}^{|\mathcal{Y}_{t-1}|\times d_{t-1}}$ is the refinement of the inherited CIL classifier, $\mathbf{F}^t_{t-1}\in\mathbb{R}^{|\mathcal{Y}_{t-1}|\times \delta_t}$ enables the use of the new features $\mathbf{\Phi}_{t}(\mathbf{x})$ for the classification into the old classes $\mathcal{Y}_{t-1},$ $\mathbf{F}^{t-1}_t\in\mathbb{R}^{(|\mathcal{Y}_t|-|\mathcal{Y}_{t-1}|)\times d_{t-1}}$ enables the use of the old features $\mathbf{\Phi}^{CIL}_{t-1}(\mathbf{x})$ for classification into the new classes $\mathcal{Y}_{t},$ and $\mathbf{M}_t \in\mathbb{R}^{|\mathcal{Y}_t|\times \delta_t} $ the classification matrix for the new classes based on the new features. It is worth noting that this presentation is mostly to highlight the connection to parameter inheritance. After initialization with (\ref{eq:expansionCIL}),
$\mathbf{M}^{CIL}_t$ can simply be updated as a full matrix.

\noindent\textbf{Inheritance in taxonomic classifier: } \label{sec:inhtax}
Parameter inheritance is a popular approach to encode the coarse-to-fine granularity of the classes defined by a taxonomic tree $\cal H$~\cite{multidetect,hierprior,deeprtc}. A single feature extractor $\boldsymbol \Phi(\mathbf{x})$ is shared by all nodes of the tree, and the taxonomic classifier (TC) implemented with~(\ref{eq:pi}), where a class is associated with each leaf node, whose 
parameter vector is a row of $\mathbf{M}$. Parameter inheritance is implemented by using as parameter vector of node ${\cal N}_i$
\begin{equation}
    \mathbf{m}_i = \mathbf{v}_i + \mathbf{m}_{{\cal P}({\cal N}_i)} \label{eq:rowi}
\end{equation}
where ${\cal P}({\cal N})$ is the parent node of ${\cal N}$. A parameter vector of $\bf 0$ is assigned to the root node of the tree. In this way, the structure of the classifier guarantees a coarse-to-fine partition of the shared feature space. When a new level of the tree is introduced, the children classes are defined incrementally over their parent super-classes. This encourages the partition of the feature space region assigned to the super-class into a set of fine-grained cells. The parameters vectors of a taxonomic classifier thus have the form
\begin{equation}
    \mathbf{m}_i = \mathbf{v}_i + \sum_{j \in {\cal A}(\mathcal{N}_i)} \mathbf{v}_j
\end{equation}
where ${\cal A}(\mathcal{N}_i)$ is the set of ancestors of ${\cal N}_i$.

\noindent\textbf{Incremental taxonomic classifier: }
The fact that parameter inheritance is central to the arguably most popular approaches to both network expansion and taxonomic classification, suggests its use for the solution of the TCIL problem. Under the TCIL task definition of Figure~\ref{fig:tree_expand}, task ${\cal T}_t$ has a taxonomic tree ${\cal H}_t$ given by~(\ref{eq:Hi}) and label set given by~(\ref{eq:yi}). As shown in Figure~\ref{fig:tree_expand}, consecutive trees ${\cal H}_{t-1}$ and ${\cal H}_t$ differ by the replacement of the oldest node (${\cal N}_t$) in ${\cal H}_{t-1}$ by its children ${\cal C}({\cal N}_t)$. Let $\mathbf{M}^{TC}_{t-1}$ be the classifier matrix of the taxonomic classifier of tree ${\cal H}_{t-1}$. Under the inheritance relation of~(\ref{eq:rowi}),  $\mathbf{M}^{TC}_{t}$ is obtained from $\mathbf{M}^{TC}_{t-1}$  by the sequence of operations:  1) eliminate the first row (node ${\cal N}_t$), 2) append $|{\cal C}({\cal N}_t)|$ rows with copies of this first row (children nodes), and 3) add to each of these rows a perturbation parameter $\mathbf{v}_i$ (to implement (\ref{eq:rowi})). These operations can be implemented with the inheritance relationship 
\begin{equation}
    \mathbf{M}^{TC}_t = \mathbf{I}_{t}^{TC} \mathbf{M}^{TC}_{t-1} + \mathbf{R}_t^{TC}
    \label{eq:Mtupdate}
\end{equation}
where 
\begin{equation}\label{eq:TCIL_PI}
    {\bf I}^{TC}_{t} = \left[\begin{array}{c}
    \mathbf{\Sigma}_{|{\cal Y}_{t-1}|} \\
    \mathbf{\Psi}_{|{\cal C}({\cal N}_t)|} \\
  \end{array} \right] \quad  
    \mathbf{R}_t^{TC}  = \left[\begin{array}{c}
    \mathbf{0}_{(|{\mathcal{Y}_{t-1}}|-1) \times d} \\
    \mathbf{V}_t
  \end{array} \right]
\end{equation}
$\mathbf{\Sigma}_k$ is as defined in (\ref{eq:inheritance}), $\mathbf{\Psi}_{k} = [\mathbf{1}_{k \times 1}, \, \mathbf{0}_{k \times (|{\cal Y}|_{t-1}-1)}]$, $\mathbf{1}_{k \times 1}$ is a vector of all ones, and $\mathbf{V}_t \in \mathbb{R}^{|{\cal C}({\cal N}_t)| \times d}$ the matrix of children perturbations. In incremental taxonomic classification only $\mathbf{V}_t$ is learned for task ${\cal T}_t$.

\begin{figure} \RawFloats
\begin{center}
\includegraphics[width=\linewidth]{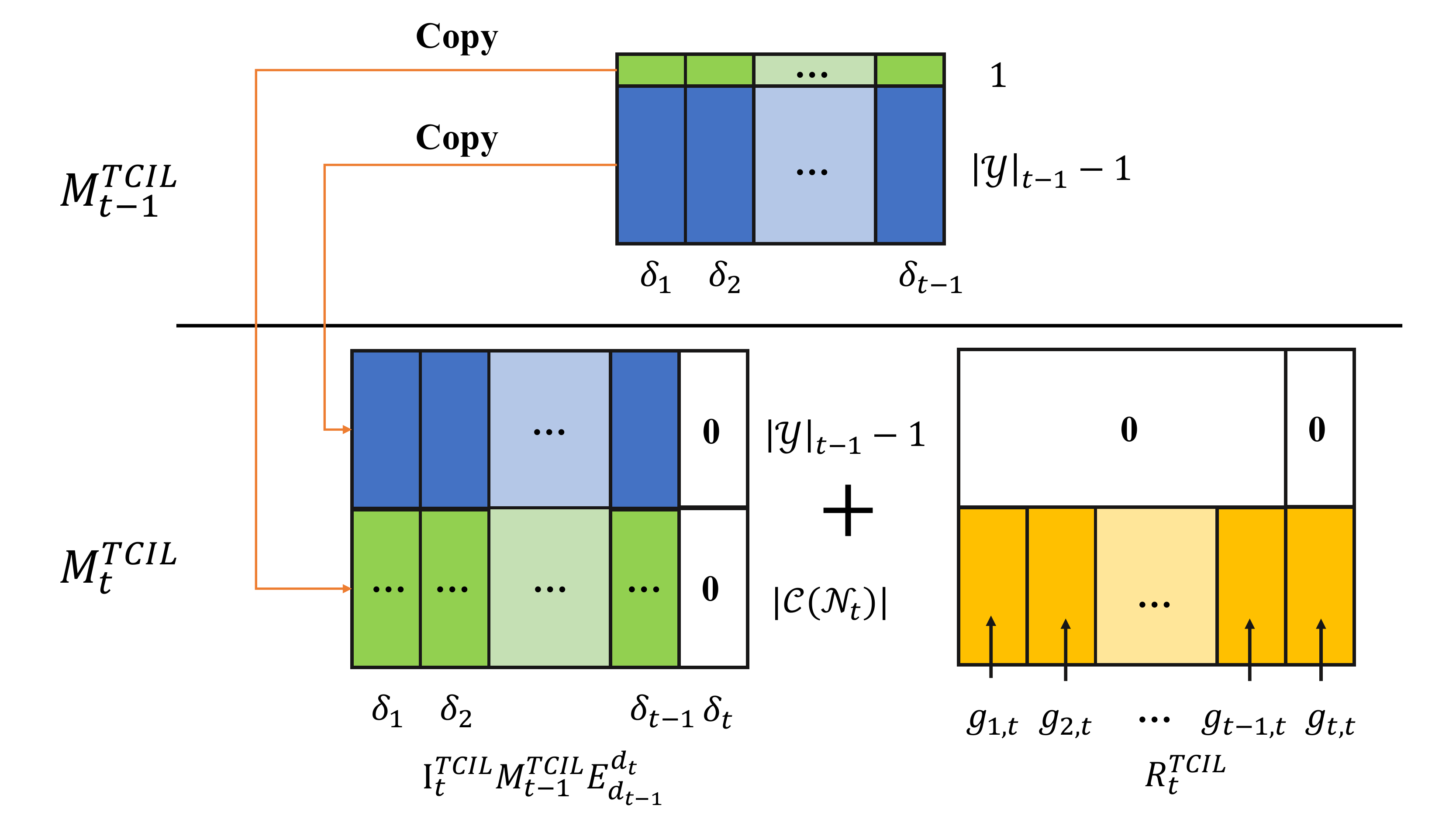}
\end{center}
\caption{Classifier expansion. Inheritance is achieved by 1) copying the first row (green part) of $\mathbf{M}^{TCIL}_{t-1}$ to the last $|\mathcal{C}(\mathcal{N}_t)|$ rows of $\mathbf{M}^{TCIL}_{t}$, 2) copying the rest $|\mathcal{Y}_{t-1}|-1$ rows (blue part) of $\mathbf{M}^{TCIL}_{t-1}$ to the first $|\mathcal{Y}_{t-1}|-1$ rows of $\mathbf{M}^{TCIL}_t$, 3) padding $\delta_t$ columns of zero to the right of$\mathbf{M}^{TCIL}_t$ and 4) adding a refinement matrix (orange part) $\mathbf{R}_t^{TCIL}$.}
\label{fig:cls_expand}
\end{figure}

\noindent\textbf{Taxonomic CIL: } \label{sec:taxonomic_cil}

So far, we have considered the incremental design of a taxonomic classifier in a static feature space. 
Taxonomic CIL (TCIL) combines an incremental taxonomic classifier and network expansion, to support a dynamic feature space. 
It builds an incremental taxonomic classifier, adding a feature extractor $\mathbf{\Phi}_i({\mathbf x})$ per task.
This simply requires the introduction of the expansion operator of (\ref{eq:EO}), as is done in standard network expansion.
Denoting by $\mathbf{M}_t^{TCIL}$, the TCIL classifier of task ${\cal T}_t$, this is implemented as
\begin{equation}
    \mathbf{M}^{TCIL}_t = \mathbf{I}_{t}^{TC} \mathbf{M}^{TCIL}_{t-1} \mathbf{E}_{d_{t-1}}^{d_t}+ \mathbf{R}_t^{TCIL}
    \label{eq:Mtcilupdate}
\end{equation}
where $\mathbf{I}_{t}^{TC}$ is defined as in (\ref{eq:TCIL_PI}), 
\begin{equation}\label{eq:TCIL_R}
    \mathbf{R}_t^{TCIL}  = \left[\begin{array}{c}
    \mathbf{0}_{(|{\cal Y}|_{t-1}-1) \times d_t} \\
    \mathbf{V}_t
  \end{array} \right]
\end{equation}
with $\mathbf{V}_t \in  \mathbb{R}^{|{\cal C}({\cal N}_t)| \times d_t}.$ During the learning of task ${\cal T}_t$ only the parameters in
$\mathbf{V}_t$ are updated, as shown in Figure~\ref{fig:cls_expand}.

\begin{figure} \RawFloats
\begin{center}
\includegraphics[width=\linewidth]{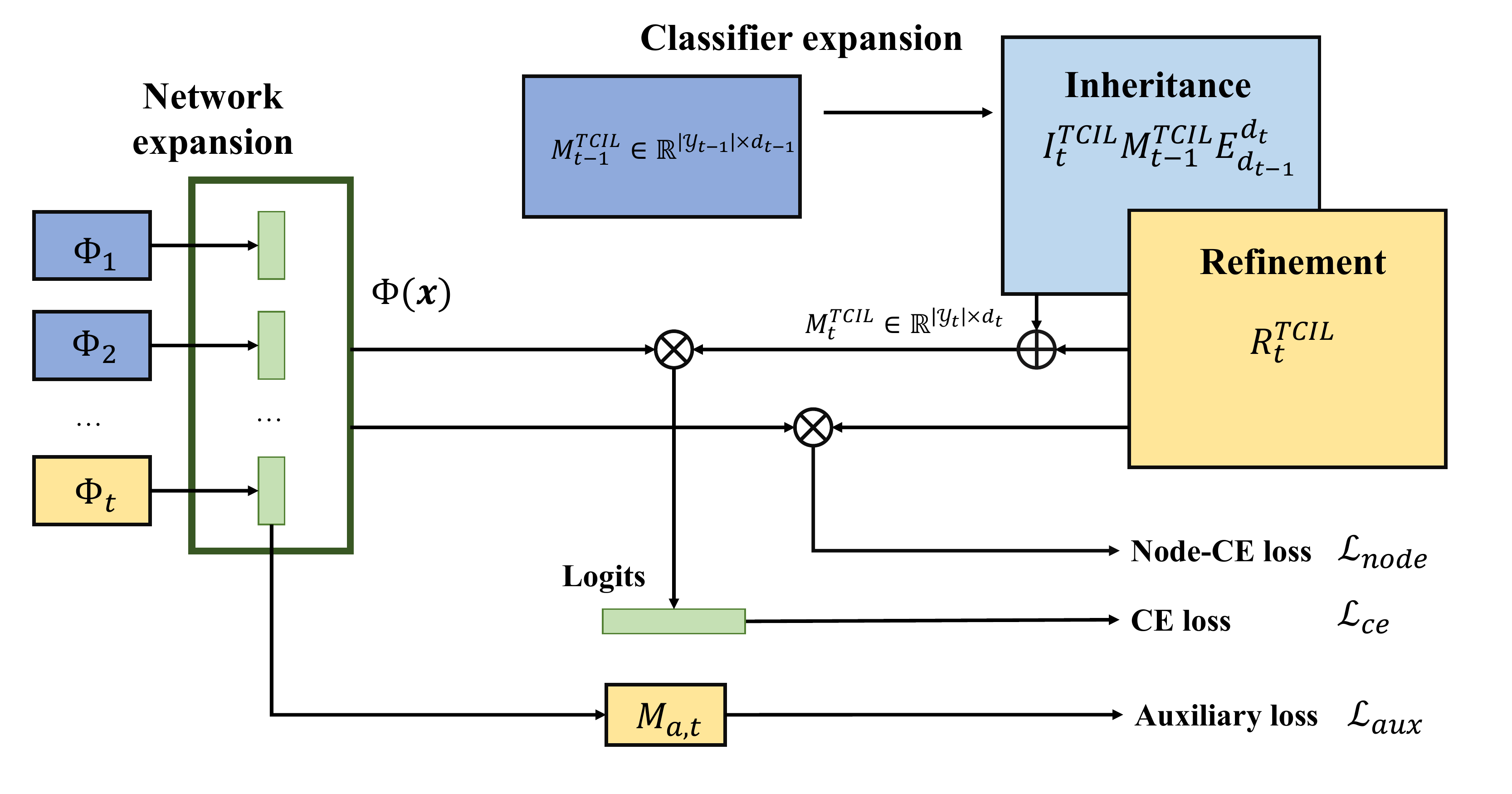}
\caption{TCIL network expansion architecture. At step $t$, the feature extractor is expanded by $\Phi_t$ and the classifier parameters $\mathbf{M}^{TCIL}_t$ assembled by inheritance and refinement. Refinement parameters and  $\Phi_t$ are trained (yellow), while previous feature extractors and inherited parameters are frozen (blue). The task is learned with several losses. An auxiliary classifier $\mathbf{M}_{a,t}$ in introduced specifically to enhance the features of the task and discarded after it.} \label{fig:main_arch}
\end{center}
\end{figure}
 
 The fact that only last $|\mathcal{Y}_t|$ rows of $\mathbf{R}_t^{TCIL}$ are non-zero implies that only the children nodes have parameter components along the dimensions of the feature space $\mathbf{\Phi}_t^{CIL}(\mathbf{x})$ corresponding to the feature extractor $\mathbf{\Phi}_t(\mathbf{x})$ added to the model by task ${\cal T}_t$. Hence, as tasks are introduced, the model gradually performs a {\it feature space expansion\/}, guaranteeing a {\it doubly hierarchical\/} classifier structure. On one hand, the parameters are organized from coarse to fine as the tree is descended, as discussed in Section \ref{sec:inhtax}. On the other, the children nodes classes expand into feature spaces orthogonal to those of their ancestors. This expansion guarantees that the model "never runs out of space" to add new fine-grained classes. This is the main advantage over the incremental taxonomic classifier of (\ref{eq:Mtupdate}), which also guarantees a coarse-to-fine class hierarchy but lacks this property.

 The gradual feature expansion of TCIL also enables control of the feature relationships by introducing a {\it hierarchical feature selection\/} vector  
 \begin{equation}
    \mathbf{g}_t = [g_{1,t}, \dots, g_{t,t}]^T \in \{0,1\}^t
\end{equation}
that controls which feature extractors $\mathbf{\Phi}_i({\bf x})$ of $\mathbf{\Phi}_t^{CIL}(\mathbf{x})$ are used by task ${\cal T}_t$.
For this it suffices  to redefine 
\begin{equation}\label{eq:TCIL_PI_g}
    \mathbf{R}_t^{TCIL} =\left[\begin{array}{c}
    \mathbf{0}_{|{\cal Y}|_{t-1} \times d_t} \\
    \mathbf{G}_t \odot  \mathbf{V}_t
  \end{array} \right]
\end{equation}
where $\odot$ is a entry-wise multiplication and 
 \begin{equation}
    \mathbf{G}_t = [g_{1,t} \mathbf{1}_{|\mathcal{Y}_t| \times \delta_1}, \; \dots, \; g_{t,t} \mathbf{1}_{|\mathcal{Y}_t| \times \delta_t}].
\end{equation}
In this way the classifier of task ${\cal T}_t$ only uses the features $\mathbf{\Phi}_i({\bf x})$ if $g_{i,t} = 1$. 

In this work we mainly adopt the {\it hierarchical feature selection strategy\/}, by making the feature selector reflect the hierarchical relationships of the taxonomic tree ${\cal H}_t$ with
\begin{equation}\label{eq:control}
    g_{i,t} = \begin{cases}
        1, \quad \text{if } \mathcal{N}_i \in (\mathcal{A}_t \cup \mathcal{N}_t) \\
        0, \quad \text{otherwise},
    \end{cases} 
\end{equation}
where $\mathcal{A}_t$ are the ancestor nodes of $\mathcal{N}_t$. In this case, the classifier of node ${\cal N}_t$ only uses feature space $\mathbf{\Phi}_i(\mathbf{x})$ if node $\mathcal{N}_i$ is an ancestor of $\mathcal{N}_t$.

Figure~\ref{fig:main_arch} summarizes the TCIL process. At step $t$, the feature extractor is expanded by $\mathbf{\Phi}_t$ and the classifier parameters $\mathbf{M}^{TCIL}_t$ assembled by inheritance and refinement, according to (\ref{eq:Mtcilupdate}). Refinement parameters and  $\Phi_t$ are trained (yellow), while previous feature extractors and inherited parameters are frozen (blue). Following DER~\cite{der}, an auxiliary classifier $\mathbf{M}_{a,t}$ in introduced specifically to enhance the features of the task and discarded after it. The task is learned with several losses, which are discussed next.

\noindent\textbf{Training objectives: } \label{loss_function}
Consider a labeled example $(\mathbf{x}, y)$ for the training of task  task $\mathcal{T}_t$. TCIL is trained with three loss functions, derived from the cross-entropy loss
$\text{CE}(\mathbf{\ell}(\mathbf{x}), y) = -\log \rho_y(\ell(\mathbf{x}))$, where $\ell(\mathbf{x}) \in\mathbb{R}^{|\mathcal{Y}_t|}$ is 
a vector of logits.

\noindent\textbf{CE Loss:} This is CE loss over leaf nodes of the tree ${\cal H}_t$,
\begin{equation}
    \mathcal{L}_{ce} = \text{CE} \Big(\mathbf{M}_{t}^{TCIL} \mathbf{\Phi}^{CIL}_t(\mathbf{x}), y \Big).
\end{equation}

\noindent\textbf{Node-CE loss:} As shown in Figure~\ref{fig:cls_expand}, task ${\cal T}_t$ replaces node ${\cal N}_t$ by its
children ${\cal C}({\cal N}_t)$. Hence $|\mathcal{Y}_t|=|\mathcal{Y}_{t-1}|-1+|\mathcal{C}(\mathcal{N}_t)|$ and the logits can be divided into
\begin{equation}
    \mathbf{\ell}(\mathbf{x}) = [\mathbf{\ell}(\mathbf{x})^{prev}\quad \mathbf{\ell}(\mathbf{x})^{curr}]
\end{equation}
where the first $|\mathcal{Y}_{t-1}|-1$ elements $\mathbf{\ell}(\mathbf{x})^{prev}\in\mathbb{R}^{|\mathcal{Y}_{t-1}|-1}$ correspond to the prediction of old classes and the last $|\mathcal{C}(\mathcal{N}_t)|$ elements $\mathbf{\ell}(\mathbf{x})^{curr}\in\mathbb{R}^{|\mathcal{C}(\mathcal{N}_t)|}$ correspond to new classes. The Node-CE loss 
\begin{equation}
    \mathcal{L}_{node} = \text{CE}\Big(\mathbf{\ell}(\mathbf{x})^{curr}, y - |\mathcal{Y}_{t-1}|+1\Big)
\end{equation}
restricts the classification to the new classes, and is only used for examples sampled from the dataset $\mathcal{D}_t$ of task ${\cal T}_t$.

\noindent\textbf{Auxiliary Loss:} Similarly to DER~\cite{der}, an auxiliary classifier $\mathbf{M}_{a,t}$ is used per task to encourage the TCIL model to learn features distinct from those learned in previous tasks. This regards all old classes (${\cal Y}_{t-1}$) as a super-class, and the new classes individually.
The label $y$ of $\mathbf{x}$ is first translated into a label
\begin{equation}
    \text{Aux}(y)= \begin{cases}
        y-|\mathcal{Y}_{t-1}|+1, \quad \mathcal{N}_y \in \mathcal{C}(\mathcal{N}_t) \\
        0, \quad \text{otherwise}
    \end{cases}
\end{equation}
for this auxiliary task, and the classifier trained with loss
\begin{equation}
    \mathcal{L}_{aux} = \text{CE} \Big(\mathbf{M}_{a,t} \mathbf{\Phi}_t(\mathbf{x}), \text{Aux}(y)\Big),
\end{equation}
We set $\mathcal{L}_{aux} = 0$ for the first task. The overall TCIL loss is
\begin{equation}
    \mathcal{L} = \mathcal{L}_{node} + \lambda_{ce} \mathcal{L}_{ce} + \lambda_{aux} \mathcal{L}_{aux}
\end{equation}
where $\lambda_{ce}$ and $\lambda_{aux}$ are hyper parameters.

\section{Experiments}

In this section, we discuss several experiments performed to evaluate TCIL.

\noindent\textbf{Baselines:}
TCIL is compared to various CIL methods. \textbf{iCaRL}~\cite{icarl} and \textbf{PODNet}~\cite{podnet} are distillation based methods, using either logits or an intermediate feature vector to perform the distillation. \textbf{Dytox}~\cite{dytox} is a recently proposed transformer-based model which also relies on distillation. \textbf{DER}~\cite{der} is a network expansion method, it learns an individual network per task. \textbf{FOSTER}~\cite{foster} further distills the old network and new networks into a compact model to reduce the increase of model size with the number of tasks.

\noindent\textbf{Experimental setup:}
All experiments are conducted on CIFAR ~\cite{cifar} and ImageNet~\cite{imagenet}. CIFAR-100 contains 50,000 training images from 100 classes, and a 2-layer taxononomy. This has 20 coarse-level super-classes and 5 fine-level classes per super-class. Following this taxonomy, we define one coarse-level task ${\cal T}_1$ including the 20 super-classes and 20 fine-level tasks $\{{\cal T}_i\}_{i=2}^{21}$ of 5 classes each.
ImageNet follows a taxonomy from WordNet~\cite{wordnet}. To build a simple benchmark, we manually select 100 classes to form the ImageNet-ILSVRC2012 dataset. Each class has 1300 images. These 100 classes are organized into a 3-layer taxonomy, with 4 coarse-level classes in the first layer, each of which has 5 descendent classes, resulting in 20 coarse-level classes in the second layer. Finally, each of these 20 classes has 5 fine-level children classes. This hierarchical structure defines 5 coarse-level (1 in the first layer and 4 in the second) tasks and 20 fine-level tasks. A more detailed discussion is included in the supplementary. For  training, the model first learns the coarse-level tasks $\mathcal{T}_1, \dots, \mathcal{T}_{N_c}$, and then the fine-level tasks $\mathcal{T}_{N_c+1}, \dots, \mathcal{T}_{N_c+N_f}$. $N_c$ and $N_f$ are the number of coarse-level and fine-level tasks. 

All methods are evaluated with several metrics. Let the accuracy of leaf node classification 
after training task ${\cal T}_i$ be $A_i$. \textit{Last Top-1 Accuracy} (Acc@1)$=A_{N_c+N_f}$ is the final performance of the model. \textit{Average Incremental Accuracy} (Avg. Acc)$=\frac{1}{N_f}\sum_{i=N_c+1}^{N_c+N_f}A_i$ evaluates the average performance over the fine-grained task sequence. This accounts for the fact that some baselines are not applicable to coarse-level tasks.

\noindent\textbf{Implementation details:}
All methods use a ResNet18 \cite{resnet} backbone. TCIL uses the hierarchical feature selection strategy of (\ref{eq:TCIL_PI_g}) and the the Decouple \cite{decouple} approach to reduce classifier for unbalanced training. Specifically, the whole model is first trained with $\mathcal{D}_t\cup\mathcal{B}$, then the feature extractor $\mathbf{\Phi}$ frozen and the training $\mathcal{D}_t\cup\mathcal{B}$ down-sampled to a class-balanced one, $\hat{\mathcal{D}_t}\cup\mathcal{B}$, on which the classifier $\mathbf{M}_t$ is tuned.
On both CIFAR-100 and ImageNet-100, the memory buffer size $|\mathcal{B}|$ is 2000. The memory buffer is built as in~\cite{icarl}. The weights of CE loss and Auxiliary loss are $\lambda_{ce} = \lambda_{aux} = 1$. On CIFAR-100 the sample rate is set as $r=0.3$. On ImageNet-100, the sample rates are $r_1=0.2$ and $r_2=0.3$. See supplementary for more details.

\noindent\textbf{Results:} Several experiments are proposed to evaluate the different components of hierarchical taxonomic. These leverage the fact that hierarchical taxonomic information can be introduced in either the training of the classifier, the design of the classifier architecture, or both. In what follows, we refer to a classifier that has no taxonomic architecture, i.e. an unconstrained parameter matrix $\mathbf{M}$, as {\it flat}, and  a classifier with taxonomic structure, i.e. that satisfies (\ref{eq:rowi}) as {\it hierarchical}. We note that, in the incremental learning setting, even a flat classifier can benefit from taxonomic information. For example, by defining early tasks as the learning of super-classes and the later tasks as the learning of fine-grained classes. We thus differentiate between the classifier and the {\it curriculum\/} used for learning, i.e. the definition of the sequence of tasks ${\cal T}_i$.  

\noindent{\bf The role of the curriculum:} These experiments use only a flat classifier.
We consider two possibilities for the curriculum. A {\bf random curriculum} groups classes into tasks randomly. This is the standard CIL setting. A {\bf semantic curriculum} groups classes by their semantic similarity. The model first learns the fine-level tasks $\mathcal{T}_{N_c+1}, \dots, \mathcal{T}_{N_c+N_f}$ defined by the taxonomy. There is still no explicit information about super-classes, and the classifier is never trained on super classes, but the class grouping of classes is derived from the taxonomy. 

Tables \ref{tab:cil_wtcil_cifar} and \ref{tab:cil_wtcil_imagenet} analyze the impact of the curriculum for a flat classifier, implemented with different CIL approaches. It is clear that most methods benefit from the semantic curriculum, on both datasets and the gains can be quite significant. For example, three of the four methods improve Acc@1 by nearly 3 points on CIFAR-100 and all methods improve Acc@1 on ImageNet-100. For the SOTA DER approach, the gains are of 3.16 on CIFAR-100 and 1.08 on ImageNet-100. These results are surprising, given that the classifier has no hierarchical structure per se, and it is never provided with information about taxonomies, super-classes, coarse-grained vs fine-grained or any related meta-data. In fact, the classifiers never perform coarse-grained classification. It appears that the semantic class organization facilitates the incremental learning of the feature spaces $\mathbf{\Phi}(\mathbf{x})$. Forcing the classifier to learn a sequence of fine-grained discrimination tasks leads to better overall features than learning a sequence of random tasks. This is consistent with the prevalence of semantic curriculum for human teaching. It also raises the question of whether the standard CIL setting (random curriculum) is the most suitable for CIL research.      
 
\begin{figure}\RawFloats
\begin{center}
\footnotesize
\begin{tabular}{| l | c | c | c | c | c | c | c | c | c |}
\hline
Metric & \multicolumn{2}{|c|}{Acc@1 (\%)} & \multicolumn{2}{|c|}{Avg. Acc (\%)}  \\
\hline
Curriculum & {Random} & {Semantic} &  {Random} & {Semantic} \\
\hline\hline
iCaRL & 41.65 & \textbf{44.12} & 59.10 & \textbf{62.15} \\
FOSTER & \textbf{48.53} & 47.59 & 60.92 & \textbf{61.44}\\
DyTox & 57.63 & \textbf{60.03} & \textbf{72.46} & 71.54\\
DER & 60.79 & \textbf{63.95} & 73.56 & \textbf{73.94} \\
\hline
\end{tabular}
\end{center}
\captionof{table}{Impact of the curriculum on CIL performance on CIFAR-100. All methods use a flat classifier.} \label{tab:cil_wtcil_cifar}
\end{figure}

\begin{figure}\RawFloats
\begin{center}
\footnotesize
\begin{tabular}{| l | c | c | c | c | c | c | c | c | c |}
\hline
Metric & \multicolumn{2}{|c|}{Acc@1 (\%)} & \multicolumn{2}{|c|}{Avg. Acc (\%)}  \\
\hline
Curriculum & {Random} & {Semantic} &  {Random} & {Semantic} \\
\hline\hline
iCaRL & 36.86 & \textbf{37.46} & 56.37 & \textbf{58.07} \\
FOSTER & 53.56 & \textbf{57.06} & 64.84 & \textbf{66.17}\\
DyTox & 61.68 & \textbf{63.88} & \textbf{72.46} & 71.98\\
DER & 67.78 & \textbf{68.86} & 76.33 & \textbf{76.50} \\
\hline
\end{tabular}
\end{center}
\captionof{table}{Impact of the curriculum on CIL performance on ImageNet-100. All methods use a flat classifier.} \label{tab:cil_wtcil_imagenet}
\end{figure}

\noindent{\bf The role of the classifier:} These experiments investigate the advantages of hierarchical classifier designs.
Both classifiers are trained with a {\bf taxonomic\/} curriculum, where tasks are introduced from coarse to fine-grained, as shown in Figure~\ref{fig:cls_expand}. The classifiers differ in the way they enforce hierarchical parameter inheritance.
CIL methods learn flat classifiers that rely only on (\ref{eq:expansionCIL}). They impose no hierarchical constraints other than those due to the curriculum. Coarse-grained tasks are learned first, and the super-class parameters used to initialize those of the later and fine-grained tasks, as discussed in Section~\ref{sec:taxonomic_cil}. TCIL learns a hierarchical classifier, which explicitly enforces the hierarchical constraints of~(\ref{eq:rowi}) through the inheritance scheme of~(\ref{eq:Mtcilupdate}).

Table~\ref{tab:classifier} compares the performance of TCIL to various CIL approaches from the literature.  First, comparing Table~\ref{tab:classifier} to Tables \ref{tab:cil_wtcil_cifar} and \ref{tab:cil_wtcil_imagenet}, shows that the flat classifier does not seem to benefit from the taxonomic curriculum. 

For most CIL approaches, the performance is inferior to that of the CIL methods that use the semantic curriculum. For example, the performance of DyTox degrades from $60.03\%$ to $58.53\%$ on CIFAR-100 and from $63.88\%$ to $61.63\%$ on ImageNet-100. This is probably due to the lack of enforcement of the hierarchical constraints. Since  CIL uses parameter inheritance just for initialization of later task classifiers, the knowledge acquired from the early coarse-grained tasks is forgotten by the time their coarse-grained descendants are introduced. Second, the explicit enforcement of hierarchical constraints by TCIL eliminates this problem. For both datasets, TCIL outperforms the best CIL method, DER, with non-trivial gains of $2.1\%$ on CIFAR-100 and $5.74\%$ on ImageNet-100. The TCIL results are also superior to  all others in Tables~\ref{tab:cil_wtcil_cifar}-\ref{tab:cil_wtcil_imagenet} showing that incremental learning benefits from the combination of a taxonomic curriculum and a taxonomic classifier.

\begin{figure}\RawFloats
\begin{center}
\footnotesize
\begin{tabular}{| l | c | c | c | c |}
\hline

Dataset & \multicolumn{2}{|c|}{CIFAR-100} & \multicolumn{2}{|c|}{ImageNet-100}  \\
\hline
Metric & {Acc@1 (\%)} & {Avg. Acc (\%)} & {Acc@1 (\%)} & Avg. Acc (\%) \\
\hline\hline
iCaRL & 41.58 & 59.65 & 37.41 & 56.92 \\ 
FOSTER & 43.80 & 57.42 & 55.42 & 62.77\\
DyTox & 58.53 & 70.87 & 61.63 & 71.56\\
DER & 66.80 & 75.30 & 67.66 & 78.21 \\
\hline
TCIL &  \textbf{68.9} &  \textbf{76.0} &  \textbf{73.4} &  \textbf{80.8}\\
\hline
\end{tabular}
\end{center}
\captionof{table}{Influence of hierarchical classifier with all of the methods use strong hierarchical curriculum.} \label{tab:classifier}
\end{figure}

\section{Conclusion}
In this paper we propose TCIL, formulating the problem and the definition of tasks and datasets. We then propose a deep learning architecture for TCIL, based on network expansion technique, which is complemented with a parameter inheritance mechanism suitable for TCIL. Results on CIFAR-100 and ImageNet-100 show that 1) organizing classes into semantic tasks helps CIL learning, 2) the implemenation of TCIL with a hierarchical classifier and a taxonomic curriculum achieves the best performance, outperforming approaches derived from SOTA CIL methods. This leads to the conclusion that appropriate introduction and utilization of taxonomic information significantly improves the quality of incremental learning.


\noindent\textbf{\Large Appendix}
\appendix


\section{Implementation details} \label{sec:implementation}

\subsection{TCIL model}

As discussed in the main text, for each dataset, tasks are divided into $N_c$ coarse-level tasks and $N_f$ fine-level tasks. Following the breadth-first search strategy, the model is first trained on the coarse-level tasks and then on the fine-level ones.

Following the decouple method of DER~\cite{der}, each task $\mathcal{T}_t, t> 1,$ is trained in two steps. In the \textit{training} stage, the whole model is updated using all available data $\mathcal{D}_t \cup \mathcal{B}$. In the \textit{decoupling} stage, the feature extractor $\mathbf{\Phi}_t(\cdot)$ is frozen and only the classifier is updated. A balanced dataset is sampled from $\mathcal{D}_t \cup \mathcal{B}$ in the second stage, to ensure that each of old and new classes has equal amount of data. This prevents the classifier from being biased towards the new classes, for which there is a lot more data.

On the CIFAR-100 dataset, for each task, training uses 170 epochs with learning rate of 0.1. The learning rate has decay of 0.1 with milestones at epochs 100 and 120. The decoupling stage is trained for 50 epochs with learning rate of 0.05. The learning rate has decay of 0.1 with milestones at epochs 15 and 30. A weight decay of 0.0005 is also used in both stages and the sample rate is set at $r=0.3$.

On ImageNet-100 dataset, for each task, training uses 220 epochs with learning rate of 0.1 and sample rates $r_1=0.2$ and $r_2=0.3$. The learning rate has decay of 0.1 with milestones at epochs 60, 120, 160 and 180. The decoupling stage is trained for 30 epochs with learning rate of 0.1. The learning rate has decay of 0.1 with milestones at epoch 15. A weight decay of 0.0005 is also used in both stages. 


\subsection{Baseline models} \label{sec:baseline}

Since TCIL leverages coarse-level tasks, which CIL models typically do not use, training was slightly adjusted to make all comparisons fair. Models without network expansion, e.g., iCaRL~\cite{icarl} or PODNet~\cite{podnet}, are trained on the coarse-level tasks suing their their training strategies. This can be seen as an initialization. For models with network expansion, such as DER, the coarse-level features learned in the first $N_c$ tasks are concatenated with the subsequent features. Fine-level tasks are then trained in the standard manner, using the original parameters.

\section{Ablation study}

In this section, we consider the influence of different factors in our experiments. 

\subsection{Feature selection strategy}
As mentioned in the main text, TCIL can leverage a feature selection strategy, which is defined by the control vector $\mathbf{g}_t$. In the main text, we introduced the \textit{hierarchical selection} strategy, where
\begin{equation}\label{eq:control_hier}
    g_{i,t} = \begin{cases}
        1, \quad \text{if } \mathcal{N}_i \in (\mathcal{A}_t \cup \mathcal{N}_t) \\
        0, \quad \text{otherwise},
    \end{cases} 
\end{equation}
$\mathcal{N}_t$ is the node expanded by task $\mathcal{T}_t$, and $\mathcal{A}_t$ are the ancestor nodes of $\mathcal{N}_t$. We consider two additional strategies. The \textit{full feature} strategy sets all components of $\mathbf{g}_t$ to one,
\begin{equation}
    g_{i,t} = 1, \quad \forall i=1,\dots, t.
\end{equation}
The \textit{orthogonal space} strategy uses only the current feature set per task
\begin{equation}\label{eq:control_orth}
    g_{i,t} = \begin{cases}
        1, \quad \text{if } i=t \\
        0, \quad \text{otherwise},
    \end{cases} 
\end{equation}
leading to an expansion into a space orthogonal to those populated by previous tasks.

Note that the \textit{orthogonal space} strategy makes the refinement matrix $\mathbf{R}_t^{TCIL}$ non-zero only in the bottom right corner (last $|\mathcal{C}(\mathcal{N}_t)|$ rows and last $\delta_t$ columns), and recursively creates a block diagonal classification matrix 
 \begin{equation}
  \mathbf{M}_t^{TCIL} = 
  \left[\begin{array}{ccccc}
  \mathbf{M}_{1,t}^{TCIL} & 0 & \dots & 0 \\
  0 & \mathbf{M}_{2,t}^{TCIL} & \dots & 0 \\
  \vdots & \vdots & \ddots & \vdots \\
  0 & 0 & \dots & \mathbf{M}_{t,t}^{TCIL} 
  \end{array} \right].
  \label{eq:inheritance}
\end{equation}
Only $\mathbf{M}_{t,t}^{TCIL}$ is trained during task $t$. This is equivalent to an orthogonal feature space expansion, where each sub-classifier $\mathbf{M}_{i,t}^{TCIL}$ is only applied to the $i$-th segment of the features $\mathbf{\Phi}_i(\mathbf{x})$. In other words, classifier $\mathbf{M}_{t,t}^{TCIL}$ only operates on the feature extractor $\mathbf{\Phi}_t(\mathbf{x})$.
Table \ref{tab:strategies} shows that the \textit{hierarchical selection} performs the best on both datasets, but the difference is small. This implies that TCIL is quite robust to the choice of features.

\begin{figure}\RawFloats
\begin{center}
\footnotesize
\begin{tabular}{| l | c | c | c | c | c | c | c | c | c |}
\hline
Dataset & \multicolumn{2}{|c|}{CIFAR-100} & \multicolumn{2}{|c|}{ImageNet-100}  \\
\hline
Metric (\%) & {Acc@1} & {Avg. Acc} &  {Acc@1} & {Avg. Acc} \\
\hline\hline
\textit{full} & {68.0} & 74.9 & 72.8 & {80.2}\\
\textit{hier. sel.} & 68.9 & {76.0} & 73.4 & {80.8} \\
\textit{orth. exp.} & 68.4 & {74.7} & {72.4} & 79.5\\
\hline
\end{tabular}
\end{center}
\captionof{table}{Impact of feature selection strategy on performance.} \label{tab:strategies}
\end{figure}

\subsection{Incremental taxonomic classifier}

In section 4 of the main text, we have introduced the incremental taxonomic classifier. This is a hierarchical CIL classifier with inheritance but no feature expansion. Recall that its inheritance and refinement matrices are
\begin{equation}\label{eq:TCIL_PI}
    {\bf I}^{TC}_{t} = \left[\begin{array}{c}
    \mathbf{\Sigma}_{|{\cal Y}_{t-1}|} \\
    \mathbf{\Psi}_{|{\cal C}({\cal N}_t)|} \\
  \end{array} \right] \quad  
    \mathbf{R}_t^{TC}  = \left[\begin{array}{c}
    \mathbf{0}_{(|{\mathcal{Y}_{t-1}}|-1) \times d} \\
    \mathbf{V}_t
  \end{array} \right]
\end{equation}

This strategy differs from traditional CIL, which uses a simple flat classifier, in the same way TCIL differs from DER. Overall this allows four options for the implementation of CIL: whether network expansion is used or not, and whether the classifier is hierarchal or flat. We rely on iCaRL~\cite{icarl} as a baseline for models without network expansion, and DER for models with the latter. A comparison of the four methods is presented in table \ref{tab:icarl_tc}. Note that in this table, all models leverage coarse-level tasks, as mentioned in \ref{sec:baseline}. The table shows that feature expansion improves accuracy by more than 20\%. However, the complexity (FLOPs) increases dramatically as well. For both feature settings, however, the use of a hierarchical classifier has a gain of 1.5\% to 2\% in the final accuracy, with a smaller increase in the average accuracy. We thus conclude that the hierarchical classifier is effective for both feature settings, either with or without feature expansion.

\begin{figure}\RawFloats
\begin{center}
\footnotesize
\begin{tabular}{| l | c  c | c | c | c | c | c | c |}
\hline
Metric & NE & HC & {Acc@1)} & {Avg. Acc} &  {FLOPs}\\
 &  &  & (\%) & (\%) &  (G)\\
\hline\hline
iCaRL & & & {41.6} & 59.7 & 1.12 \\
iCaRL + HC & & \checkmark& 43.2 & {60.8} & 1.12 \\
DER & \checkmark & & 66.8 & {75.3} & 23.4\\
TCIL & \checkmark & \checkmark & 68.9 & {76.0} & 23.4\\
\hline
\end{tabular}
\end{center}
\captionof{table}{Comparison of model performances with/without network expansion (NE) and hierarchical classifier (HC), on CIFAR-100. Note that the original iCaRL model has final accuracy of 44.1\% on CIFAR-100 when no coarse-level tasks are leveraged. The addition of coarse-level tasks decreases the performance of models without feature expansion. This has already been discussed in the main article.} \label{tab:icarl_tc}
\end{figure}

\subsection{Feature dimensions}
\begin{figure} \RawFloats
\begin{center}
    \includegraphics[width=\linewidth]{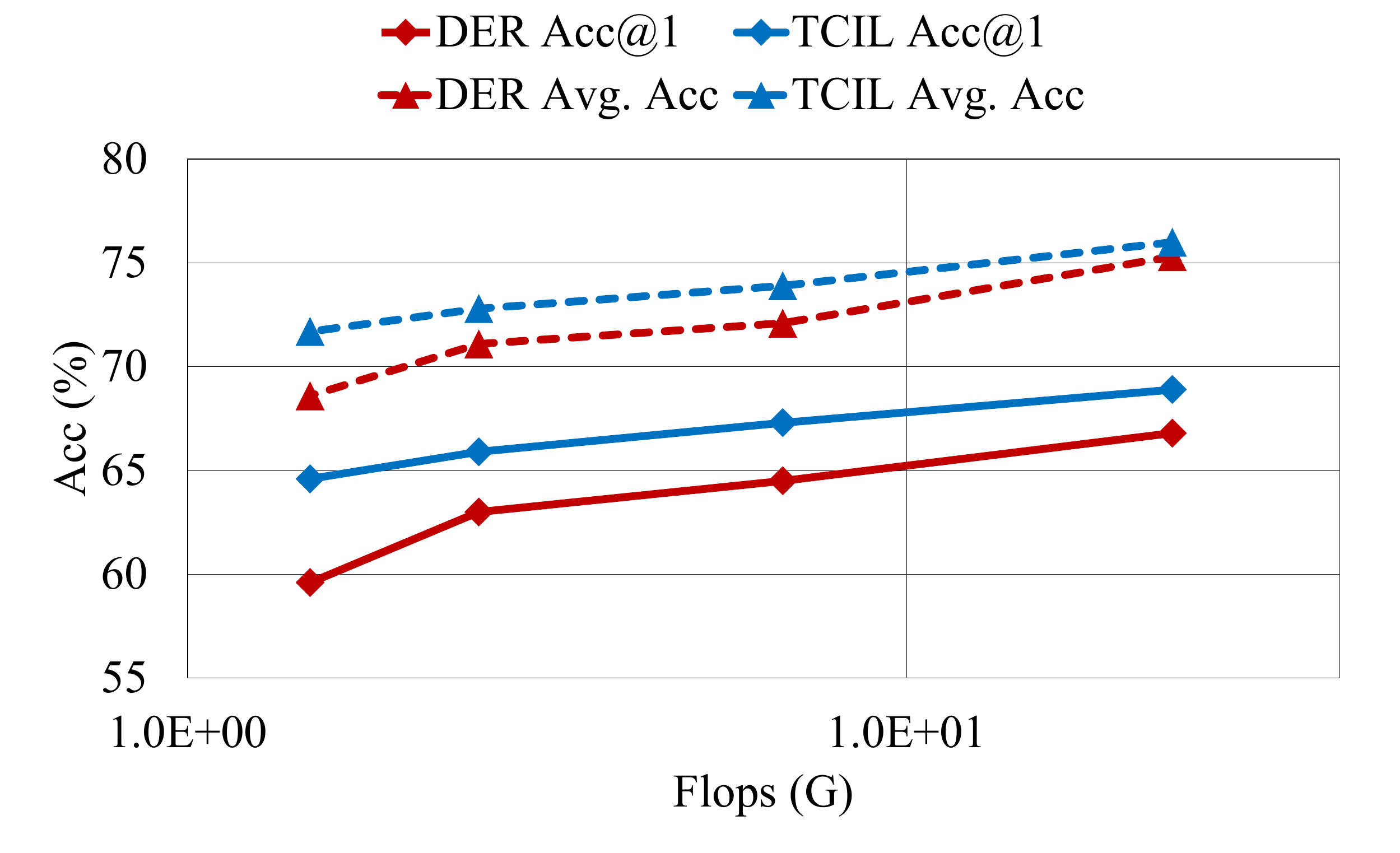}
    \caption{Accuracy vs. FLOPs of DER and TCIL models on CIFAR-100.} \label{fig:feature_dim}
\end{center}
\end{figure}

In the main text we defined a feature size of $\delta_t = 512$ per task ${\cal T}_t$. In this section, we explore different settings for $\delta_t$, so as to significantly reduce model FLOPs. Table \ref{tab:feature_dim} and figure \ref{fig:feature_dim} show that both models lose accuracy as the feature size decreases. However, the decay is smaller for TCIL, which always outperforms DER for a given FLOPs. This indicates that the benefits of TCIL hold even for models of smaller scale. In fact, its gains are larger for these models (5.8\% at 1.48G FLOPs vs 1.8\% at 23.4G FLOPs). Note that the TCIL model of 1.48G FLOPS is competitive with the DER model of 6.72G FLOPS. 

\begin{figure}\RawFloats
\begin{center}
\footnotesize
\begin{tabular}{| l | c | c | c | c | c | c |}
\hline
Metric & \multicolumn{2}{|c|}{Acc@1 (\%)} & \multicolumn{2}{|c|}{Avg. Acc (\%)}  & \multicolumn{2}{|c|}{FLOPs (G)} \\
\hline
Method & {DER} & {TCIL} & {DER} & {TCIL} & {DER} & {TCIL}\\
\hline\hline
(512, 512) & {66.8} & 68.9 & {75.3} & 76.0 & 23.4 & 23.4\\
(512, 256) & 64.5 & {67.3} & {72.1} & 73.9 & 6.72 & 6.72\\
(512, 128) & 63.0 & {65.9} & {71.1} & 72.8 & 2.54 & 2.54\\
(512, 64) & 59.6 & {64.4} & {68.6} & 71.7 & 1.48 & 1.48\\
\hline
\end{tabular}
\end{center}
\captionof{table}{Impact of feature sizes on CIFAR-100.  The left-most column shows the values of $\delta_i$ used per taxonomic layer. For example, (512, 256) means that we use $\delta = 512$ for tasks whose class nodes have depth 1 and $\delta = 256$ for those of depth of 2.} \label{tab:feature_dim}
\end{figure}

\subsection{Tree structure}
In this section, we explore the effect of the taxonomy tree structure on the performance of TCIL and DER models. As explained in \ref{sec:baseline}, DER model training leverages coarse-level tasks. All experiments are performed on ImageNet-100.

\subsubsection {Tree depth}
We start by exploring the influence of tree depth. As mentioned in the main paper, we use a default tree of 3 layers, with 4 nodes in the first taxonomic level, 5 child nodes per parent in the second taxonomic level, and 5 in the third (bottom) level. We denote this as a $4\times 5 \times 5$ tree. 

In this section, we consider trees of 2 and 4 layers and experiment with the corresponding task settings. We consider a 2-layer tree setup as $20\times 5$, and a 4-layer tree 
 setup as $2\times 2 \times 5 \times 5$. Under each setting we still have $N_f=20$. However, for coarse-level tasks, we have $N_c=1$ for the 2-layer tree (similar to CIFAR-100), $N_c=5$ for the 3-layer tree (default), and $N_c=7$ for the 4-layer tree. We continue to use breadth-first tree traversal.

Table \ref{tab:tree_depths} shows TCIL outperforms DER for all tree depths, with a gain of over 2\% in all cases. This shows that TCIL is robust to the curriculum used to introduce classes. However, the preformance degrades with tree depth for both models. This may be because deeper trees require more data to train coarse-level nodes, leaving only a smaller amount of data to train the fine-level nodes that are difficult to classify. Furter experiments, with larger datasets, are needed to clarify this issue.

\begin{figure}\RawFloats
\begin{center}
\footnotesize

\begin{tabular}{| l | c | c | c | c | c | c |}
\hline
Metric & \multicolumn{2}{|c|}{Acc@1 (\%)} & \multicolumn{2}{|c|}{Avg. Acc (\%)}  \\
\hline
Method & {DER} & {TCIL} & {DER} & {TCIL}\\
\hline\hline
2-layer & 72.5 & 74.6 & 81.9 & 81.6 \\
3-layer & 68.7 & 73.4 & 79.1 & 80.8 \\
4-layer & 67.0 & 70.3 & 77.0 &77.8 \\
\hline
\end{tabular}
\end{center}
\captionof{table}{Comparison of different tree depths, on ImageNet-100. Sample rate settings are: $r=0.3$ for 2-layer tree; $r_1=0.2, r_2=0.3$ for 3-layer tree; $r_1=0.15, r_2=0.15, r_3=0.3$ for the 4-layer tree.} \label{tab:tree_depths}
\end{figure}

\subsubsection {Tree expansion order}

We next study how the tree traversal method impacts performance. So far, we have assumed a BFS strategy, which first visits the coarsest level nodes, then moves down one layer to visit all nodes on the second level, and so on. We now consider two other traversal methods: \textit{depth-first search} (DFS) and Random traversal. The former first recursively visits the first child node of current node until it reaches the leaf nodes, then moves on to the second child node, etc. until the entire tree has been traversed. The latter consists of randomly choosing nodes from all expandable (leaf nodes) nodes in the current tree until all leaf nodes have been visited. 

Table \ref{tab:traverse} compares the results of the different tree traversal strategies. Note that the Random results are the average performance over three randomly generated traversal strategies. The table shows that TCIL has a gain of over 2\% in final accuracy for DER, for both DFS and random traversal. For TCIL, performance is higher for BFS, but similar for the three traversal strategies. This shows that TCIL is robust to the task introduction order and to imbalance of the sub-trees generated during training. This suggests that it should also fare well for applications where the entire tree is imbalanced, but this remains to be tested.

\begin{figure}\RawFloats
\begin{center}
\footnotesize
\begin{tabular}{| l | c | c | c | c | c | c |}
\hline
Metric & \multicolumn{2}{|c|}{Acc@1 (\%)} & \multicolumn{2}{|c|}{Avg. Acc (\%)}  \\
\hline
Method & {DER} & {TCIL} & {DER} & {TCIL}\\
\hline\hline
3-layer-BFS & 68.7 & 73.4 & 79.1 & 80.8 \\
3-layer-DFS & 69.0 & 72.4 & 79.8 & 80.1 \\
3-layer-Random & 69.8 & 71.9 & 75.3 & 74.5 \\
\hline
\end{tabular}
\end{center}
\captionof{table}{Comparison of different expansion orders, on ImageNet-100.} \label{tab:traverse}
\end{figure}

{\small
\bibliographystyle{ieee_fullname}
\bibliography{egbib}

\begin{thebibliography}{10}\itemsep=-1pt

\bibitem{expnet}
Karim Ahmed, Mohammad~Haris Baig, and Lorenzo Torresani.
\newblock Network of experts for large-scale image categorization.
\newblock In {\em Computer Vision--ECCV 2016: 14th European Conference,
  Amsterdam, The Netherlands, October 11--14, 2016, Proceedings, Part VII 14},
  pages 516--532. Springer, 2016.

\bibitem{expert}
Rahaf Aljundi, Punarjay Chakravarty, and Tinne Tuytelaars.
\newblock Expert gate: Lifelong learning with a network of experts.
\newblock In {\em Proceedings of the IEEE Conference on Computer Vision and
  Pattern Recognition}, pages 3366--3375, 2017.

\bibitem{survey3}
Matthias De~Lange, Rahaf Aljundi, Marc Masana, Sarah Parisot, Xu Jia,
  Ale{\v{s}} Leonardis, Gregory Slabaugh, and Tinne Tuytelaars.
\newblock A continual learning survey: Defying forgetting in classification
  tasks.
\newblock {\em IEEE transactions on pattern analysis and machine intelligence},
  44(7):3366--3385, 2021.

\bibitem{imagenet}
Jia Deng, Wei Dong, Richard Socher, Li-Jia Li, Kai Li, and Li Fei-Fei.
\newblock Imagenet: A large-scale hierarchical image database.
\newblock In {\em 2009 IEEE conference on computer vision and pattern
  recognition}, pages 248--255. Ieee, 2009.

\bibitem{mnist}
Li Deng.
\newblock The mnist database of handwritten digit images for machine learning
  research.
\newblock {\em IEEE Signal Processing Magazine}, 29(6):141--142, 2012.

\bibitem{podnet}
Arthur Douillard, Matthieu Cord, Charles Ollion, Thomas Robert, and Eduardo
  Valle.
\newblock Podnet: Pooled outputs distillation for small-tasks incremental
  learning.
\newblock In {\em Computer Vision--ECCV 2020: 16th European Conference,
  Glasgow, UK, August 23--28, 2020, Proceedings, Part XX 16}, pages 86--102.
  Springer, 2020.

\bibitem{dytox}
Arthur Douillard, Alexandre Ram{\'e}, Guillaume Couairon, and Matthieu Cord.
\newblock Dytox: Transformers for continual learning with dynamic token
  expansion.
\newblock In {\em Proceedings of the IEEE/CVF Conference on Computer Vision and
  Pattern Recognition}, pages 9285--9295, 2022.

\bibitem{tnn}
Wonjoon Goo, Juyong Kim, Gunhee Kim, and Sung~Ju Hwang.
\newblock Taxonomy-regularized semantic deep convolutional neural networks.
\newblock In {\em Computer Vision--ECCV 2016: 14th European Conference,
  Amsterdam, The Netherlands, October 11-14, 2016, Proceedings, Part II 14},
  pages 86--101. Springer, 2016.

\bibitem{resnet}
Kaiming He, Xiangyu Zhang, Shaoqing Ren, and Jian Sun.
\newblock Deep residual learning for image recognition.
\newblock In {\em Proceedings of the IEEE conference on computer vision and
  pattern recognition}, pages 770--778, 2016.

\bibitem{distillation}
Geoffrey Hinton, Oriol Vinyals, and Jeff Dean.
\newblock Distilling the knowledge in a neural network.
\newblock {\em arXiv preprint arXiv:1503.02531}, 2015.

\bibitem{decouple}
Bingyi Kang, Saining Xie, Marcus Rohrbach, Zhicheng Yan, Albert Gordo, Jiashi
  Feng, and Yannis Kalantidis.
\newblock Decoupling representation and classifier for long-tailed recognition.
\newblock {\em arXiv preprint arXiv:1910.09217}, 2019.

\bibitem{wsn}
Haeyong Kang, Rusty John~Lloyd Mina, Sultan Rizky~Hikmawan Madjid, Jaehong
  Yoon, Mark Hasegawa-Johnson, Sung~Ju Hwang, and Chang~D Yoo.
\newblock Forget-free continual learning with winning subnetworks.
\newblock In {\em International Conference on Machine Learning}, pages
  10734--10750. PMLR, 2022.

\bibitem{afc}
Minsoo Kang, Jaeyoo Park, and Bohyung Han.
\newblock Class-incremental learning by knowledge distillation with adaptive
  feature consolidation.
\newblock In {\em Proceedings of the IEEE/CVF conference on computer vision and
  pattern recognition}, pages 16071--16080, 2022.

\bibitem{ewc}
James Kirkpatrick, Razvan Pascanu, Neil Rabinowitz, Joel Veness, Guillaume
  Desjardins, Andrei~A Rusu, Kieran Milan, John Quan, Tiago Ramalho, Agnieszka
  Grabska-Barwinska, et~al.
\newblock Overcoming catastrophic forgetting in neural networks.
\newblock {\em Proceedings of the national academy of sciences},
  114(13):3521--3526, 2017.

\bibitem{adns}
Yajing Kong, Liu Liu, Zhen Wang, and Dacheng Tao.
\newblock Balancing stability and plasticity through advanced null space in
  continual learning.
\newblock In {\em Computer Vision--ECCV 2022: 17th European Conference, Tel
  Aviv, Israel, October 23--27, 2022, Proceedings, Part XXVI}, pages 219--236.
  Springer, 2022.

\bibitem{cifar}
Alex Krizhevsky, Geoffrey Hinton, et~al.
\newblock Learning multiple layers of features from tiny images.
\newblock 2009.

\bibitem{lwf}
Zhizhong Li and Derek Hoiem.
\newblock Learning without forgetting.
\newblock {\em IEEE transactions on pattern analysis and machine intelligence},
  40(12):2935--2947, 2017.

\bibitem{gem}
David Lopez-Paz and Marc'Aurelio Ranzato.
\newblock Gradient episodic memory for continual learning.
\newblock {\em Advances in neural information processing systems}, 30, 2017.

\bibitem{piggy}
Arun Mallya, Dillon Davis, and Svetlana Lazebnik.
\newblock Piggyback: Adapting a single network to multiple tasks by learning to
  mask weights.
\newblock In {\em Proceedings of the European Conference on Computer Vision
  (ECCV)}, pages 67--82, 2018.

\bibitem{survey2}
Marc Masana, Xialei Liu, Bart{\l}omiej Twardowski, Mikel Menta, Andrew~D
  Bagdanov, and Joost van~de Weijer.
\newblock Class-incremental learning: survey and performance evaluation on
  image classification.
\newblock {\em IEEE Transactions on Pattern Analysis and Machine Intelligence},
  2022.

\bibitem{wordnet}
George~A Miller.
\newblock {\em WordNet: An electronic lexical database}.
\newblock MIT press, 1998.

\bibitem{hierprior}
Radford~M Neal and Babak Shahbaba.
\newblock Improving classification when a class hierarchy is available using a
  hierarchy-based prior.
\newblock 2007.

\bibitem{fisher}
Razvan Pascanu and Yoshua Bengio.
\newblock Revisiting natural gradient for deep networks.
\newblock {\em arXiv preprint arXiv:1301.3584}, 2013.

\bibitem{survey1}
Haoxuan Qu, Hossein Rahmani, Li Xu, Bryan Williams, and Jun Liu.
\newblock Recent advances of continual learning in computer vision: An
  overview.
\newblock {\em arXiv preprint arXiv:2109.11369}, 2021.

\bibitem{rpsnet}
Jathushan Rajasegaran, Munawar Hayat, Salman~H Khan, Fahad~Shahbaz Khan, and
  Ling Shao.
\newblock Random path selection for continual learning.
\newblock {\em Advances in Neural Information Processing Systems}, 32, 2019.

\bibitem{progressprompt}
Anastasia Razdaibiedina, Yuning Mao, Rui Hou, Madian Khabsa, Mike Lewis, and
  Amjad Almahairi.
\newblock Progressive prompts: Continual learning for language models.
\newblock {\em arXiv preprint arXiv:2301.12314}, 2023.

\bibitem{icarl}
Sylvestre-Alvise Rebuffi, Alexander Kolesnikov, Georg Sperl, and Christoph~H
  Lampert.
\newblock icarl: Incremental classifier and representation learning.
\newblock In {\em Proceedings of the IEEE conference on Computer Vision and
  Pattern Recognition}, pages 2001--2010, 2017.

\bibitem{pnn}
Andrei~A Rusu, Neil~C Rabinowitz, Guillaume Desjardins, Hubert Soyer, James
  Kirkpatrick, Koray Kavukcuoglu, Razvan Pascanu, and Raia Hadsell.
\newblock Progressive neural networks.
\newblock {\em arXiv preprint arXiv:1606.04671}, 2016.

\bibitem{multidetect}
Ruslan Salakhutdinov, Antonio Torralba, and Josh Tenenbaum.
\newblock Learning to share visual appearance for multiclass object detection.
\newblock In {\em CVPR 2011}, pages 1481--1488. IEEE, 2011.

\bibitem{hier_survey}
Carlos~N Silla and Alex~A Freitas.
\newblock A survey of hierarchical classification across different application
  domains.
\newblock {\em Data Mining and Knowledge Discovery}, 22:31--72, 2011.

\bibitem{geodl}
Christian Simon, Piotr Koniusz, and Mehrtash Harandi.
\newblock On learning the geodesic path for incremental learning.
\newblock In {\em Proceedings of the IEEE/CVF conference on Computer Vision and
  Pattern Recognition}, pages 1591--1600, 2021.

\bibitem{iNat}
Grant Van~Horn, Oisin Mac~Aodha, Yang Song, Yin Cui, Chen Sun, Alex Shepard,
  Hartwig Adam, Pietro Perona, and Serge Belongie.
\newblock The inaturalist species classification and detection dataset.
\newblock In {\em Proceedings of the IEEE conference on computer vision and
  pattern recognition}, pages 8769--8778, 2018.

\bibitem{foster}
Fu-Yun Wang, Da-Wei Zhou, Han-Jia Ye, and De-Chuan Zhan.
\newblock Foster: Feature boosting and compression for class-incremental
  learning.
\newblock In {\em Computer Vision--ECCV 2022: 17th European Conference, Tel
  Aviv, Israel, October 23--27, 2022, Proceedings, Part XXV}, pages 398--414.
  Springer, 2022.

\bibitem{nscl}
Shipeng Wang, Xiaorong Li, Jian Sun, and Zongben Xu.
\newblock Training networks in null space of feature covariance for continual
  learning.
\newblock In {\em Proceedings of the IEEE/CVF conference on Computer Vision and
  Pattern Recognition}, pages 184--193, 2021.

\bibitem{deeprtc}
Tz{-}Ying Wu, Pedro Morgado, Pei Wang, Chih{-}Hui Ho, and Nuno Vasconcelos.
\newblock Solving long-tailed recognition with deep realistic taxonomic
  classifier.
\newblock {\em CoRR}, abs/2007.09898, 2020.

\bibitem{der}
Shipeng Yan, Jiangwei Xie, and Xuming He.
\newblock Der: Dynamically expandable representation for class incremental
  learning.
\newblock In {\em Proceedings of the IEEE/CVF Conference on Computer Vision and
  Pattern Recognition}, pages 3014--3023, 2021.

\bibitem{hdcnn}
Zhicheng Yan, Hao Zhang, Robinson Piramuthu, Vignesh Jagadeesh, Dennis DeCoste,
  Wei Di, and Yizhou Yu.
\newblock Hd-cnn: hierarchical deep convolutional neural networks for large
  scale visual recognition.
\newblock In {\em Proceedings of the IEEE international conference on computer
  vision}, pages 2740--2748, 2015.

\bibitem{si}
Friedemann Zenke, Ben Poole, and Surya Ganguli.
\newblock Continual learning through synaptic intelligence.
\newblock In {\em International conference on machine learning}, pages
  3987--3995. PMLR, 2017.

\bibitem{largemargin}
Jian Zheng, Chuan Luo, Tianrui Li, and Hongmei Chen.
\newblock A novel hierarchical feature selection method based on large margin
  nearest neighbor learning.
\newblock {\em Neurocomputing}, 497:1--12, 2022.

\bibitem{bcnn}
Xinqi Zhu and Michael Bain.
\newblock B-cnn: branch convolutional neural network for hierarchical
  classification.
\newblock {\em arXiv preprint arXiv:1709.09890}, 2017.

\end{thebibliography}
}

\end{document}